\pdfoutput=1

\documentclass[letterpaper, 10 pt, conference]{ieeeconf}
\usepackage[T1]{fontenc}
\usepackage{cite}
\usepackage{amssymb,amsfonts}
\usepackage{blindtext}
\makeatletter
\let\NAT@parse\undefined
\makeatother
\usepackage{algorithmic}
\usepackage{graphicx}
\usepackage{textcomp}
\usepackage{mathtools}
\usepackage{changes}
\usepackage{subcaption}
\usepackage{algorithm}
\usepackage{xcolor}
\usepackage{color, soul}
\usepackage{amsmath}
\usepackage{booktabs}
\usepackage{multirow}
\usepackage{xstring}
\usepackage{hhline}
\usepackage{siunitx}
\usepackage{comment}
\usepackage{physics}
\usepackage{tikz}
\usepackage{hyperref}
\usepackage{cleveref}

\crefformat{section}{\S#2#1#3} 
\crefformat{subsection}{\S#2#1#3}
\crefformat{subsubsection}{\S#2#1#3}

\floatname{algorithm}{Procedure}

\newcommand{\argmin}{\mathop{\mathrm{argmin}}}

\definecolor{rvc}{RGB}{0, 0, 255}
\definecolor{suyong}{RGB}{0, 255, 0}
\definecolor{comment}{RGB}{0, 0, 0}
\definecolor{cv2}{RGB}{0, 0, 0}
\definecolor{qw}{RGB}{0, 0, 0}
\definecolor{qwr}{RGB}{0, 0, 0}
\definecolor{qwe}{RGB}{0, 0, 0}

\sethlcolor{lightgray}

\makeatletter
\DeclareRobustCommand{\iscircle}{\mathord{\mathpalette\is@circle\relax}}
\newcommand\is@circle[2]{%
  \begingroup
  \sbox\z@{\raisebox{\depth}{$\m@th#1\bigcirc$}}%
  \sbox\tw@{$#1\square$}%
  \resizebox{!}{\ht\tw@}{\usebox{\z@}}%
  \endgroup
}
\makeatother

\IEEEoverridecommandlockouts                              

\DeclareUnicodeCharacter{200B}{{\hskip 0pt}}

\usepackage{caption}
\newcommand{\rom}[1]{\uppercase\expandafter{\romannumeral #1\relax}}

\title{\LARGE \bf
A Single Correspondence Is Enough: Robust Global Registration \\to Avoid Degeneracy in Urban Environments}

\author{Hyungtae Lim$^{1,3}$, \textit{Student Member, IEEE}, Suyong Yeon$^{2}$, Soohyun Ryu$^{2}$, Yonghan Lee$^{2}$, Youngji Kim$^{2}$, \\ Jaeseong Yun$^{2}$, Euigon Jung$^{2}$, Donghwan Lee$^{2*}$, and Hyun Myung$^{1*}$, \textit{Senior Member, IEEE}
\thanks{$^*$Corresponding authors: Donghwan Lee and Hyun Myung}
\thanks{$^{1}$Hyungtae Lim and $^{1}$Hyun Myung are with the School of Electrical Engineering, KI-AI at KAIST (Korea Advanced Institute of Science and Technology), Daejeon, 34141, South Korea. {\tt\small \{shapelim, hmyung\}@kaist.ac.kr} \hfill \break 
\indent $^{2}$The authors are with NAVER LABS, Seongnam-si, Gyeonggi-do, South Korea.{\tt\scriptsize \{suyong.yeon, soohyun.ryu, yh.l, youngji.b.kim, jeaseong.yun, j.eg, donghwan.lee\}@naverlabs.com} \hfill \break  
\indent This work was supported by the Institute of Information \& communications Technology Planning \& Evaluation~(IITP) grant funded by the Korea government~(MSIT) (No. 2019-0-01309, Development of AI Technology for Guidance of a Mobile Robot to its Goal with Uncertain Maps in Indoor/Outdoor Environments)​\break
\indent $^3$\textcolor{comment}{Hyungtae Lim was with NAVER LABS, Seongnam-si, Gyeonggi-do, South Korea as a research intern} and is supported by the BK21 FOUR~(Republic of Korea).}
}

\begin{document}

\captionsetup[figure]{labelformat={default},labelsep=period,name={fig.}}

\maketitle

\begin{abstract}

Global registration using 3D point clouds is a crucial technology for mobile platforms to achieve localization or manage loop-closing situations. In recent years, numerous researchers have proposed global registration methods to address a large number of outlier correspondences. Unfortunately, \textcolor{qw}{the degeneracy} problem, which represents the phenomenon in which the number of estimated inliers becomes lower than three, is still potentially inevitable. To tackle the problem, a \textcolor{cv2}{degeneracy-robust decoupling-based} global registration method is proposed, called \textit{Quatro}. In particular, our method \textcolor{cv2}{employs} \textit{quasi-SO(3) estimation} by leveraging the Atlanta world assumption in urban environments to avoid degeneracy in rotation estimation. Thus, the minimum degree of freedom (DoF) of our method is reduced from three to one. As verified in indoor and outdoor 3D LiDAR datasets, our proposed method yields robust global registration performance compared with other global registration methods, even \textcolor{qwr}{for} distant point cloud pairs. Furthermore, the experimental results confirm the applicability of our method as a coarse alignment. Our code is available: \href{https://github.com/url-kaist/quatro}{\texttt{https://github.com/url-kaist/quatro}} 

\section{Introduction} \label{sec:intro}
\vspace{-0.1cm}

\end{abstract}

\textcolor{comment}{3D point cloud registration is a method to align 3D point clouds by estimating relative pose or poses between two or more 3D point clouds}~\cite{holz2015registration}. 3D registration methods are widely utilized in various research fields  such as object recognition~\cite{ashbrook1998finding,chua19963d,cho2014projection,belongie2002shape}, mapping~\cite{kim2019gp,geiger2012kitticvpr,lim2020normal}, \textcolor{comment}{localization}~\cite{cui2015real,javanmardi2017towards,fontanelli2007fast,song21flooprplan}, and so forth. In particular, 3D registration is also a fundamental task for terrestrial mobile platforms to estimate ego-motion~\cite{behley2018efficient,zhang2014loam}, i.e. odometry, and to manage loop closing situations~\cite{kim2018scancontext,chang2020spoxelnet}.

Accordingly, the registration algorithms are mainly divided into two categories. One is the local registration \cite{besl1992method,segal2009gicp,chetverikov2002trimmed_icp,koide2020vgicp, rusinkiewicz2001efficienticp, pomerleau2013comparing}, \textcolor{qwr}{and the other is the global registration~\cite{fischler1981ransac,dong2017gh-icp,yang2015goicp,zhou2016fast,yang2020teaser}.} Iterative Closest Point (ICP) \cite{besl1992method} is a renowned local registration method and it has made an impact on subsequent studies. Unfortunately, the ICP-variants set point pairs by using a greedy, exhaustive nearest neighbor~(NN) search for every iteration. Consequently, the local registration methods are only applicable when two point clouds, namely \textcolor{qw}{source and target}, are close enough and \textcolor{qwr}{nearly} overlapped~\cite{pomerleau2013comparing}.  \textcolor{cv2}{Otherwise, the correspondences via the NN search are likely to become invalid; thus,} \textcolor{comment}{the result of the local registration} may get caught in the local minima~\cite{koide2020vgicp}. 

The global registration\textcolor{cv2}{~\cite{fischler1981ransac,dong2017gh-icp,yang2015goicp,zhou2016fast,yang2020teaser}}, on which this paper places emphasis, aims to estimate the relative pose between distant and partially overlapped point clouds. \textcolor{qw}{Because the global registration is relatively invariant to the initial pose difference}, the global registration is usually used to provide an initial alignment close enough for the local registration to allow the estimate to converge on the global minima.
The global registration can be further classified into correspondence-based\textcolor{comment}{~\cite{fischler1981ransac,zhou2016fast,yang2020teaser,tzoumas2019outlier}} and correspondence-free methods~\cite{bernreiter2021phaser,rouhani2011correspondence,brown2019family}, but the latter ones are beyond our scope.

\begin{figure}[t!]
    \captionsetup{font=footnotesize}
    \centering
	\begin{subfigure}[b]{0.48\textwidth}
		\includegraphics[width=1\textwidth]{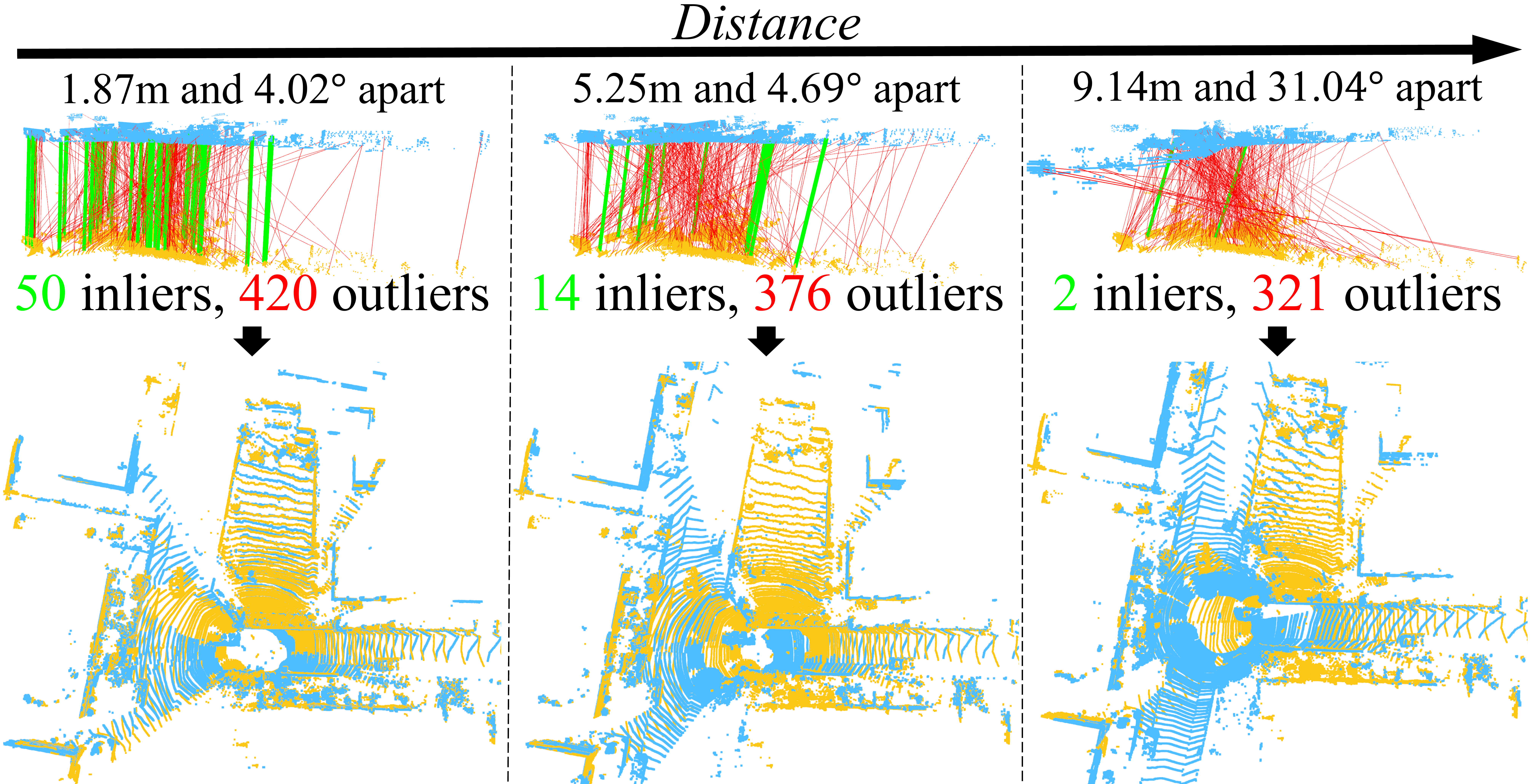}
	\end{subfigure}
	\vspace{-0.45cm}
	\caption{(T-B, L-R): Before and after the application of our proposed method called \textit{Quatro} in KITTI dataset~\cite{geiger2012kitticvpr} when two distant and partially overlapped point clouds, i.e. source (cyan) and target (yellow), are given. \textcolor{cv2}{As the distance between the two viewpoints of source and target becomes farther, it gives rise to an increase in the ratio of outliers within the putative correspondences and simultaneously reduces the number of inliers, which results in the performance degradation of correspondence-based global registration methods in general~\cite{fischler1981ransac,zhou2016fast,tzoumas2019outlier}.} Under the circumstances, our proposed method shows robust performance, overcoming the effect of outliers, as well as the degeneracy issue. The red and green lines denote outlier and inlier correspondences, respectively~(best viewed in color).}
	\label{fig:overview}
	\vspace{-0.6cm}
\end{figure}

Some of the well-known correspondence-based methods are \textcolor{qwe}{random sample consensus}~(RANSAC)~\cite{fischler1981ransac} and its variants \cite{papazov2012rigid,chum2003locally,choi1997performance,schnabel2007efficient}, yet these are known to become slow and brittle with high outlier rates ($>50$\%)~\cite{tzoumas2019outlier}. On the other hand, branch-and-bound (BnB)-based methods~\cite{olsson2008branch,pan2019multibnb,hartley2009global} have been proposed for theoretical optimality guarantees. Unfortunately, BnB-based methods are not applicable to real-world applications because they are time-consuming~($>50$~sec)~\cite{lei2017fast}.

Aiming for both outlier-robust and fast registrations, graduated non-convexity~(GNC) has been introduced based on Black-Rangarajan duality~\cite{black1996unification,zhou2016fast}. \textcolor{qw}{For instance,} Zhou \textit{et al.}~\cite{zhou2016fast} showed that GNC outperforms previous approaches and is faster \textcolor{qw}{than existing methods by at least ten times, while} tolerating 70-80\% of outliers. As a result, many researchers have employed GNC to achieve robust pose estimation~\cite{tzoumas2019outlier,yang2020teaser,sun2021iron,yang2020gnc}. In the meantime, semidefinite programming and sums of squares relaxation-based methods have been proposed to solve problems in polynomial-time with certifiable optimality guarantees~\cite{vandenberghe1996semidefinite,briales2017convex,maron2016point,carlone2016planar}. Further, novel graph-theoretic methods also have been proposed to prune spurious correspondences effectively~\cite{shi2020robin,sun2021ransic}.

\textcolor{comment}{Unfortunately, a degeneracy problem is still potentially inevitable.
Degeneracy usually refers to various phenomena in the perceptually degraded environments, such as corridors, mines, and so forth~\cite{ebadi2021dare,hinduja2019degeneracy,westman2019degeneracy}. However, in this paper, we specify degeneracy as the phenomenon in which the number of the estimated inliers becomes lower than the \textcolor{qw}{minimum number of inliers} during registration, which is \textcolor{qwr}{generally} three in the 3D space. \textcolor{qw}{Note that} this effect can occur not only in the perceptually degraded environments but also in some cases where two distant and partially overlapped point clouds are given, resulting in catastrophic failure of global registration.}


\textcolor{comment}{To sum up, the reason for the degeneracy problem is mainly twofold. First, there is a sparsity issue with 3D LiDAR sensors~\cite{lim2021patchwork,lim2021erasor,lu2019deepvcp,lim2022pago}, which means, as the distance from the sensor coordinate becomes farther away, a point cloud becomes too sparse; thus, it induces inaccurate feature matching. That is, even though the two point clouds represent the same space locally, the local areas have different densities because the source and target are captured in a different \textcolor{cv2}{location}. Consequently, the sparsity issue directly affects the quality of feature descriptors and then gives rise to more \textcolor{cv2}{false correspondences. Thus, the ratio of the outliers within the putative correspondences increases, as shown in Fig.~\ref{fig:overview}}. Second, these outlier pairs let the outlier-rejection algorithms, such as graph-theoretic pruning methods \cite{shi2020robin,sun2021ransic} or weight update in GNC~\cite{zhou2016fast,yang2020teaser}, reject the putative correspondences dramatically to prune outliers. Accordingly, too many correspondences are occasionally filtered. Finally, the number of the estimated inliers occasionally becomes less than three in the process, resulting in degeneracy.}

To tackle the degeneracy problem, we propose a novel \textcolor{qwr}{global registration method}, called \textit{Quatro}, which is a combination of the words \textit{Quasi-} and \textit{cuatro} \textcolor{qwe}{meaning 4-DoF (degree of freedom)}. Some degeneracy-aware SLAM frameworks \textcolor{qw}{have been proposed in}~\cite{ebadi2021dare,hinduja2019degeneracy,westman2019degeneracy}. However, to the best of our knowledge, this is the first attempt to deal with \textcolor{qw}{the} degeneracy in global registration by reducing the \textcolor{qwr}{minimum number of required correspondences} of SE(3) from three to one, named \textit{quasi-SE(3)} by decoupling SE(3) estimation into
quasi-SO(3) estimation followed by \textcolor{
qwr}{component-wise
translation estimation (COTE)~\cite{yang2020teaser}}.

In summary, the contribution of this paper is threefold.
\begin{itemize}
	\item \textcolor{cv2}{To avoid degeneracy, we propose a novel rotation estimation method called quasi-SO(3) estimation by utilizing the characteristics of urban environments.} 
	\item As a result, our proposed method shows a more promising performance over the state-of-the-art methods on real-world indoor/outdoor 3D LiDAR datasets.
	\item  In particular, it is remarkable that Quatro-\texttt{c2f}, a simple coarse-to-fine scheme, outperforms \textcolor{comment}{the state-of-the-art methods, including deep learning-based approaches.} Therefore, we finally confirm the suitability of our proposed method as a coarse alignment.
\end{itemize}
\section{Quatro: Quasi-SE(3) Estimation}
\vspace{-0.1cm}
\subsection{Problem Definition}

First, we begin by denoting source and target point cloud which are captured by a 3D LiDAR sensor on different \textcolor{comment}{viewpoints} as $\mathcal{P}$ and $\mathcal{Q}$, respectively. Then, let \textcolor{qw}{us define} $\mathcal{P}=\{\mathbf{p}_{1}, \mathbf{p}_{2}, \dots, \mathbf{p}_{N}\}$ and $\mathcal{Q}=\{\mathbf{q}_{1}, \mathbf{q}_{2}, \dots, \mathbf{q}_{M}\}$, where each point of the clouds, $\mathbf{p}_{i}~(1 \leq i \leq N)$ and $\mathbf{q}_j~{(1 \leq j \leq M)}$ $\in \mathbb{R}^3$, consists of $\{x, y, z\}$ in the Cartesian coordinate. It is assumed that the $xy$-plane of $\mathcal{P}$ and $\mathcal{Q}$ are already aligned with that of mobile platforms to utilize the Atlanta world assumption~\cite{straub2017manhattan}~(see Section~\rom{2}.\textit{C}).

Next, let $(i, j) \in \mathcal{A}$ be \textcolor{qwr}{candidate} correspondences \textcolor{cv2}{where $i$ and $j$ denote the indices of points in $\mathcal{P}$ and $\mathcal{Q}$, respectively.} In practice, $\mathcal{A}$ inevitably includes inherent outlier set $\mathcal{O}$, such that $\mathcal{O} \cup \mathcal{O}^{c} = \mathcal{A}$. This happens \textcolor{qwr}{in most of the} 3D point feature matching algorithms~\cite{tzoumas2019outlier, rusu2009fpfh,yew20183dfeat}. Accordingly, the relationship of each pair can be expressed as follows:

\vspace{-0.3cm}

\begin{equation}
\mathbf{q}_{j}=\mathbf{R}\mathbf{p}_{i}+\mathbf{t}+\boldsymbol{\epsilon}_{ij}
\label{eq:pt_relation}
\end{equation}

\vspace{-0.1cm}

\noindent where $\mathbf{R} \in \mathrm{SO}(3)$ and $\mathbf{t}\in\mathbb{R}^{3}$ are the \textcolor{qwe}{relative} rotation and translation, respectively, and $\boldsymbol{\epsilon}_{ij}\in\mathbb{R}^{3}$ denotes the unknown measurement noise. That is, $\boldsymbol{\epsilon}_{ij}$ is Gaussian noise if $(i,j)~\in~\mathcal{O}^{c}$ or is \textcolor{qw}{irregular error} if $(i,j)~\in~\mathcal{O}$. Finally, our objective function could be defined as follows:

\vspace{-0.4cm}

\begin{equation}
  \hat{\mathbf{R}}, \hat{\mathbf{t}} =\argmin_{\mathbf{R} \in \mathrm{SO}(3), \mathbf{t} \in \mathbb{R}^{3}}  \sum_{(i,j) \in \mathcal{A}}\rho\Big( r(\mathbf{q}_{j}-\mathbf{R} \mathbf{p}_{i}-\mathbf{t})\Big) 
  \label{eqn:goal}
\end{equation}

\vspace{-0.1cm}

\noindent where $\rho(\cdot)$ denotes surrogate function \cite{tzoumas2019outlier} to suppress undesirable large errors produced by $\mathcal{O}$ and $r(\cdot)$ denotes the squared residual function, \textcolor{qw}{i.e. $\abs{\cdot}^2$}. In summary, our goal is to \textcolor{comment}{estimate} $\hat{\mathbf{R}}$ and $\hat{\mathbf{t}}$ while suppressing the effect of outliers as much as possible \textcolor{cv2}{by employing $\rho(\cdot)$}.

\subsection{Decoupling Rotation and Translation Estimation}

According to \cite{horn1987closed} and \cite{arun1987least}, $\hat{\mathbf{R}}$ and $\hat{\mathbf{t}}$ can be easily obtained in closed form by decoupling rotation and translation estimation. To this end, first of all, two pairs ($i$, $j$) and ($i^\prime$, $j^\prime$) are subtracted to cancel out the effect of $\mathbf{t}$ by using (\ref{eq:pt_relation}). 
For simplicity, let $\boldsymbol{\alpha}_k~=~\mathbf{p}_{i^{\prime}} - \mathbf{p}_{i}$; $\boldsymbol{\beta}_k~=~\mathbf{q}_{j^{\prime}} - \mathbf{q}_{j}$; $\boldsymbol{\epsilon}_k~=~\boldsymbol{\epsilon}_{i^{\prime} j^{\prime}} - \boldsymbol{\epsilon}_{ij}$, and let $K$ be the total number of these translation invariant measurements (TIMs)~\cite{yang2020teaser}. Note that $K = \abs{\mathcal{A}}$ by subtracting consecutive pairs to build TIMs in a chain form to minimize computational cost and \textcolor{qwe}{the first TIM is made by the subtraction between the first and $\abs{\mathcal{A}}$-th pair.} 

Then, the relationship between $\boldsymbol{\alpha}_{k}$ and $\boldsymbol{\beta}_{k}$ can be expressed as $\boldsymbol{\beta}_{k}=\mathbf{R}\boldsymbol{\alpha}_{k}+\boldsymbol{\epsilon}_{k}$. Accordingly, $\hat{\mathbf{R}}$ is \textcolor{qw}{estimated} as follows:

\vspace{-0.5cm}

\begin{equation}
\hat{\mathbf{R}}=\underset{\mathbf{R} \in \mathrm{SO}(3)}{\argmin } \sum_{k=1}^{K} \min \Big( w_k r({\boldsymbol{\beta}}_{k}- \mathbf{R}{\boldsymbol{\alpha}}_{k}), \bar{c}^{2}\Big)
\label{eqn:decouple_rot}
\end{equation}

\noindent where $w_k$ denotes the weight of each pair and $\bar{c}$ denotes the \textcolor{qwr}{truncation} parameter to suppress the effect of potential outliers. This is followed by COTE\textcolor{comment}{~\cite{yang2020teaser}} to calculate $\hat{\mathbf{t}}$ in an element-wise way as follows:

\vspace{-0.2cm}

\begin{equation}
^{l}{\hat{\mathbf{t}}}=\underset{^{l}\mathbf{t} \in \mathbb{R}}{\argmin } \sum_{(i,j) \in \mathcal{A}}^{\abs{\mathcal{A}}} \min \bigg( \frac{r({^{l}\mathbf{t}}-{^{l}\mathbf{v}_{ij}})}{\sigma_{ij}^2}, {\bar{c}^{2}} \bigg)
\label{eqn:decouple_trans}
\end{equation}

\vspace{-0.2cm}

\noindent where $\mathbf{v}_{ij}=\mathbf{q}_{j}-{\hat{\mathbf{R}}} \mathbf{p}_{i}$, $\sigma_{ij}$ is the noise bound, and $^l(\cdot)$ \textcolor{qwr}{denotes} the $l$-th element of a 3D vector, where $l=1,2,3$, i.e. $\hat{\mathbf{t}}=\{{^1\hat{\mathbf{t}}},{^2\hat{\mathbf{t}}},{^3\hat{\mathbf{t}}}\}$. 

Thus, \textcolor{qwr}{(\ref{eqn:goal}) is decomposed into (\ref{eqn:decouple_rot}) followed by (\ref{eqn:decouple_trans}) using~$\hat{{\mathbf{R}}}$.}



\begin{figure}[t!]
    \begin{subfigure}[b]{0.23\textwidth}
		\includegraphics[width=1.0\textwidth]{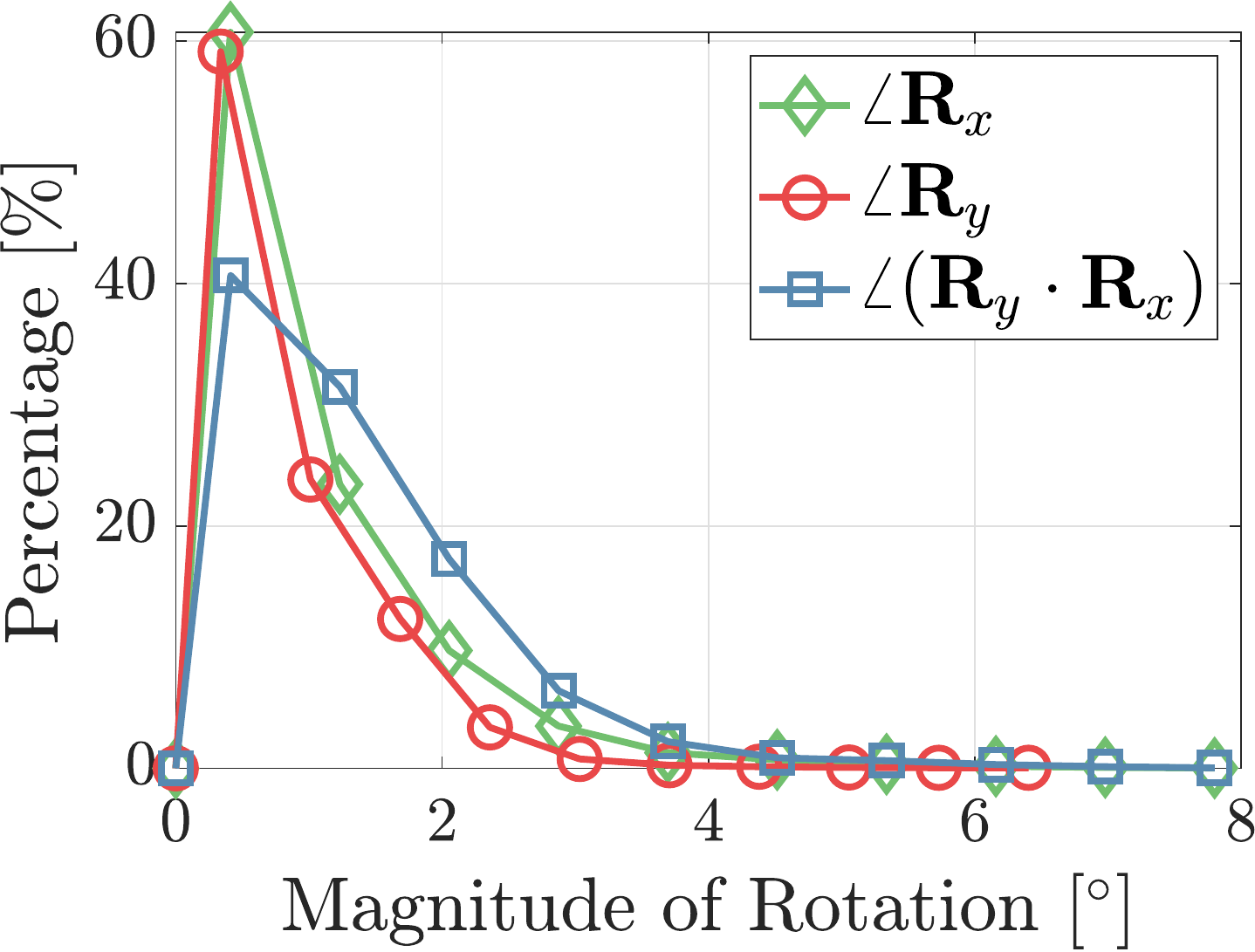}
		\caption{}
	\end{subfigure}
	\begin{subfigure}[b]{0.23\textwidth}
		\includegraphics[width=1.0\textwidth]{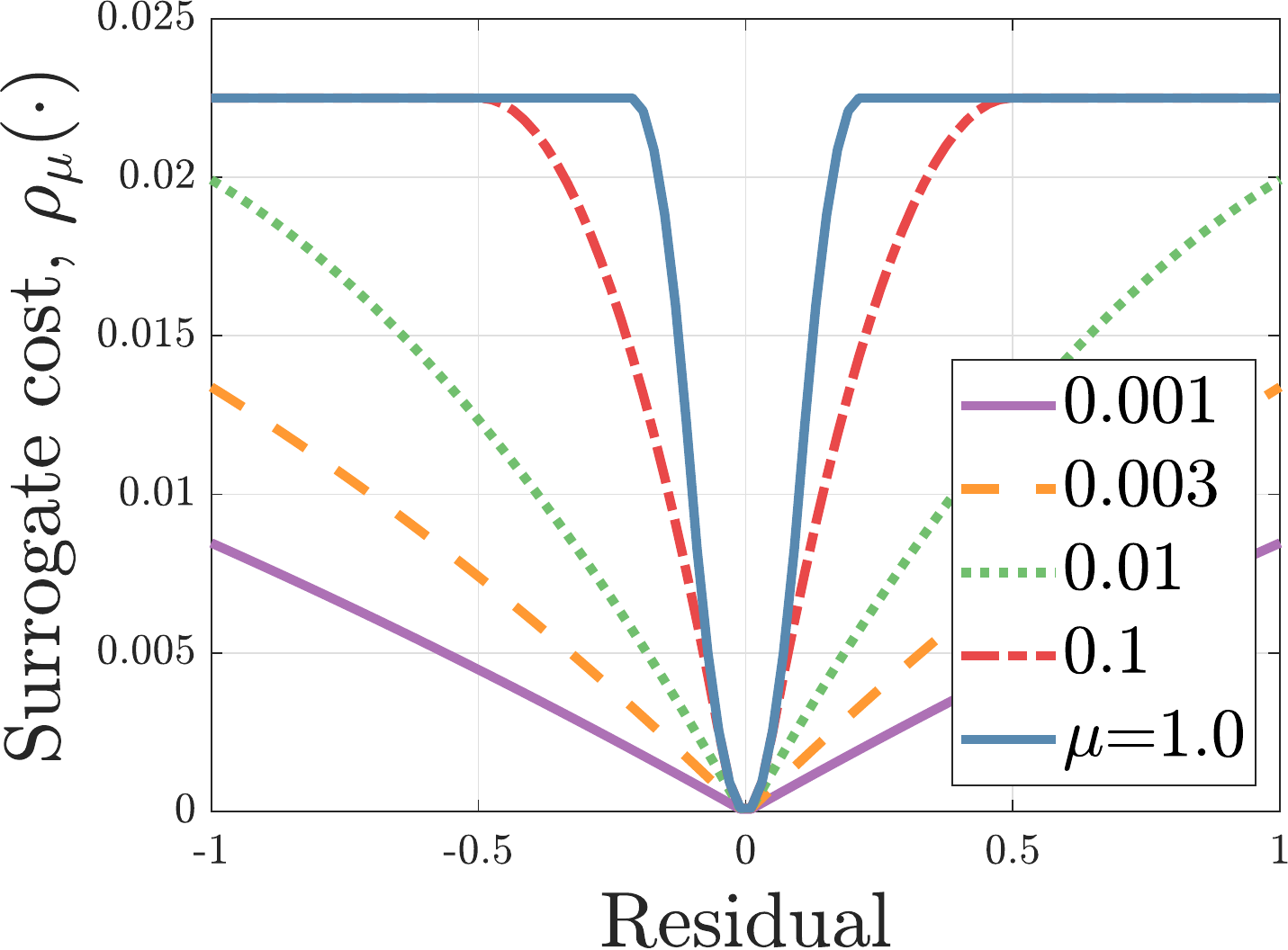}
		\caption{}
	\end{subfigure}
	\vspace{-0.2cm}
	\captionsetup{font=footnotesize}
    \caption{(a) Probability distribution function of the magnitudes of \textcolor{qwe}{relative} rotation between poses of source and target whose distances are between 0.5 to 10 m away in the KITTI dataset~\cite{geiger2012kitticvpr}. (b) Surrogate function $\rho_{\mu}(\cdot)$ that becomes convex if $\mu \rightarrow 0$ or truncated least squares if $\mu \rightarrow \infty$, where $\bar{c}=0.15$ (best viewed in color).}
	\label{fig:exp_evidences}
	\vspace{-0.6cm}
\end{figure}

\subsection{Quasi-SO(3) in Urban Environments}

 Next, we introduce the concept of \textit{quasi-SO(3)} to note that the relative rotation can be \textcolor{qwr}{approximated} as a pure yaw rotation in urban environments, which \textcolor{qwr}{was} postulated by Kim \textit{et al.} \cite{kim2019gp} and Scaramuzza \cite{scaramuzza20111dransac}. \textcolor{cv2}{This is because, even though urban environments are not flat, two poses corresponding to the source and target cloud \textcolor{qwr}{can be} locally \textcolor{qwr}{approximated} to be coplanar based on the Atlanta world assumption \cite{straub2017manhattan}. Therefore, the yaw rotation is considered to be dominant than roll and pitch rotations.}
 
 \textcolor{cv2}{To be more concrete, let us specify a \textcolor{qwe}{relative} rotation matrix $\mathbf{R}$ as $\mathbf{R}_{z} \cdot \mathbf{R}_{y} \cdot \mathbf{R}_{x}$, where $\mathbf{R}_{z}$, $\mathbf{R}_{y}$, and $\mathbf{R}_{x}$ denote rotation with respect to $z$, $y$, and $x$ \textcolor{qwr}{axes}, respectively. That is, \textcolor{qw}{these rotation elements} \textcolor{qwr}{denote} yaw, pitch, and roll, respectively. As previously mentioned, the \textcolor{qwe}{relative changes} of pitch and roll were observed to be much smaller than that of yaw in urban canyons, i.e. $\angle \mathbf{R}_{z} \gg \angle (\mathbf{R}_{y} \cdot \mathbf{R}_{x})$, where $\angle \mathbf{R}=\cos^{-1}{\frac{\Tr(\mathbf{R}) - 1}{2}}$. In addition, as shown in Fig.~\ref{fig:exp_evidences}(a), $\angle (\mathbf{R}_{y} \cdot \mathbf{R}_{x})$ is usually smaller than $10^\circ$; thus, the small angle assumption~\cite{youn2021state} is applicable, such that $\mathbf{R}_{y} \cdot \mathbf{R}_{x} \approx \mathbf{I}_3$, where $\mathbf{I}_n$ denotes \textcolor{qwr}{an} $n\times n$ identity matrix. This assumption results in $\mathbf{R} \approx \mathbf{R}_{z}$ and thus simplifies our goal to estimate $\mathbf{R}_{z}$ directly. Finally, this assumption reduces \textcolor{qwe}{DoF} of rotation from three to one, making our method robust against degeneracy. \textcolor{cv2}{For simplicity, the approximated $\mathbf{R}$ is denoted by $\mathbf{R}_{+}$.}}
 
 \textcolor{cv2}{One might argue that this assumption does not hold in non-flat regions or if two viewpoints are no longer located in coplanar regions as the distance between two viewpoints increases~\cite{shan2020lio,lee2016accurate,qin2018vins}. However, most current mapping/navigation systems employ an inertial navigation system (INS). Accordingly, roll and pitch angles are fully observable by estimating the horizontal plane from the gravity vector~\cite{qin2018vins,lim2021avoiding}. Consequently, the roll and pitch can be expressed as absolute states in the world coordinate, which means these are drift-free~\cite{hesch2014camera}. Thus, this still allows our goal to be valid, while restricting the relative rotation to relative yaw rotation (see Section~\rom{4}.\textit{D}).}
 
\subsection{Quasi-SO(3) Estimation using Graduated Non-Convexity}

To estimate $\mathbf{R}_{+}$, GNC with a truncated least square~\cite{tzoumas2019outlier} is introduced. \textcolor{comment}{For simplicity}, specifying the $k$-th measurement pair, i.e. $\boldsymbol{\alpha}_{k}$ and  $\boldsymbol{\beta}_{k}$, as $\boldsymbol{\zeta}_k$, (\ref{eqn:decouple_rot}) can be redefined as $\argmin _{\mathbf{R}_{+}} \sum_{k=1}^{K} \rho_{\mu}\left(r\left(\boldsymbol{\zeta}_k, \mathbf{R}_{+}\right)\right)$, where $\rho_{\mu}(\cdot)$ denotes surrogate function governed by parameter $\mu$, as shown in Fig.~\ref{fig:exp_evidences}(b). For this, the equation is first rewritten by leveraging Black-Rangarajan duality as follows~\cite{zhou2016fast, yang2020teaser}:

\vspace{-0.45cm}
\begin{equation}
\hat{\mathbf{R}}_{+} = \argmin _{\mathbf{R}_{+} \in \mathrm{SO}(3), w_{k} \in[0,1]} \sum_{k=1}^{K}\left[w_{k} r\left(\boldsymbol{\zeta}_k, \mathbf{R}_{+}\right)+\Phi_{\rho_{\mu}}\left(w_{k}\right)\right]
\label{eq:black_rangarajan}
\end{equation}

\vspace{-0.10cm}

\noindent where $\Phi_{\rho_{\mu}}\left(w_{k}\right)=\frac{\mu\left(1-w_{k}\right)}{\mu+w_{k}} \bar{c}^{2}$ is \textcolor{qwr}{a} penalty term \cite{yang2020gnc}. Unfortunately, the objective function cannot be directly solved~\cite{zhou2016fast}. Thus, (\ref{eq:black_rangarajan}) is solved by using alternating optimization \textcolor{qwr}{as follows}:

\vspace{-0.3cm}

\begin{equation}
\hat{\mathbf{R}}^{(t)}_{+}=\underset{\mathbf{R}_{+}}{\argmin} \sum_{i=1}^{K} \hat{w}^{(t-1)}_{k} r\left(\boldsymbol{\zeta}_k, \mathbf{R}_{+}\right),
\label{eq:rotation}
\end{equation}

\vspace{-0.45cm}

\begin{equation}
\hat{\mathbf{W}}^{(t)}=\underset{w_{k} \in[0,1]}{\argmin } \sum_{k=1}^{K}\left[w_{k} r(\boldsymbol{\zeta}_k, \hat{\mathbf{R}}^{(t)}_{+})+\Phi_{\rho_{\mu}}\left(w_{k}\right)\right]
\label{eq:weight}
\end{equation}

\noindent where the superscript $^{(t)}$ denotes the $t$-th iteration and $\hat{\mathbf{W}}=\operatorname{diag}\left(\hat{w}_{1}, \hat{w}_{2}, \ldots, \hat{w}_{K}\right)$. Each \textcolor{qwr}{$\hat{w}_{k}$} can be solved in a truncated closed form as follows:

\vspace{-0.35cm}

\begin{equation}
\hat{w}_{k}^{(t)}= \begin{cases}0 & \text { if } \tilde{r}_{k} \in\left[\frac{\mu+1}{\mu} \bar{c}^{2},+\infty\right] \\ \bar{c} \sqrt{\frac{\mu(\mu+1)}{\tilde{r}_{k}}}-\mu & \text { if } \tilde{r}_{k} \in\left[\frac{\mu}{\mu+1} \bar{c}^{2}, \frac{\mu+1}{\mu} \bar{c}^{2}\right] \\ 1 & \text { if } \tilde{r}_{k} \in\left[0, \frac{\mu}{\mu+1} \bar{c}^{2}\right]\end{cases}
\label{eq:truncated_weight}
\end{equation}

\begin{figure}[t!]
    \captionsetup{font=footnotesize}
	\centering
    \def\svgwidth{0.46\textwidth}
    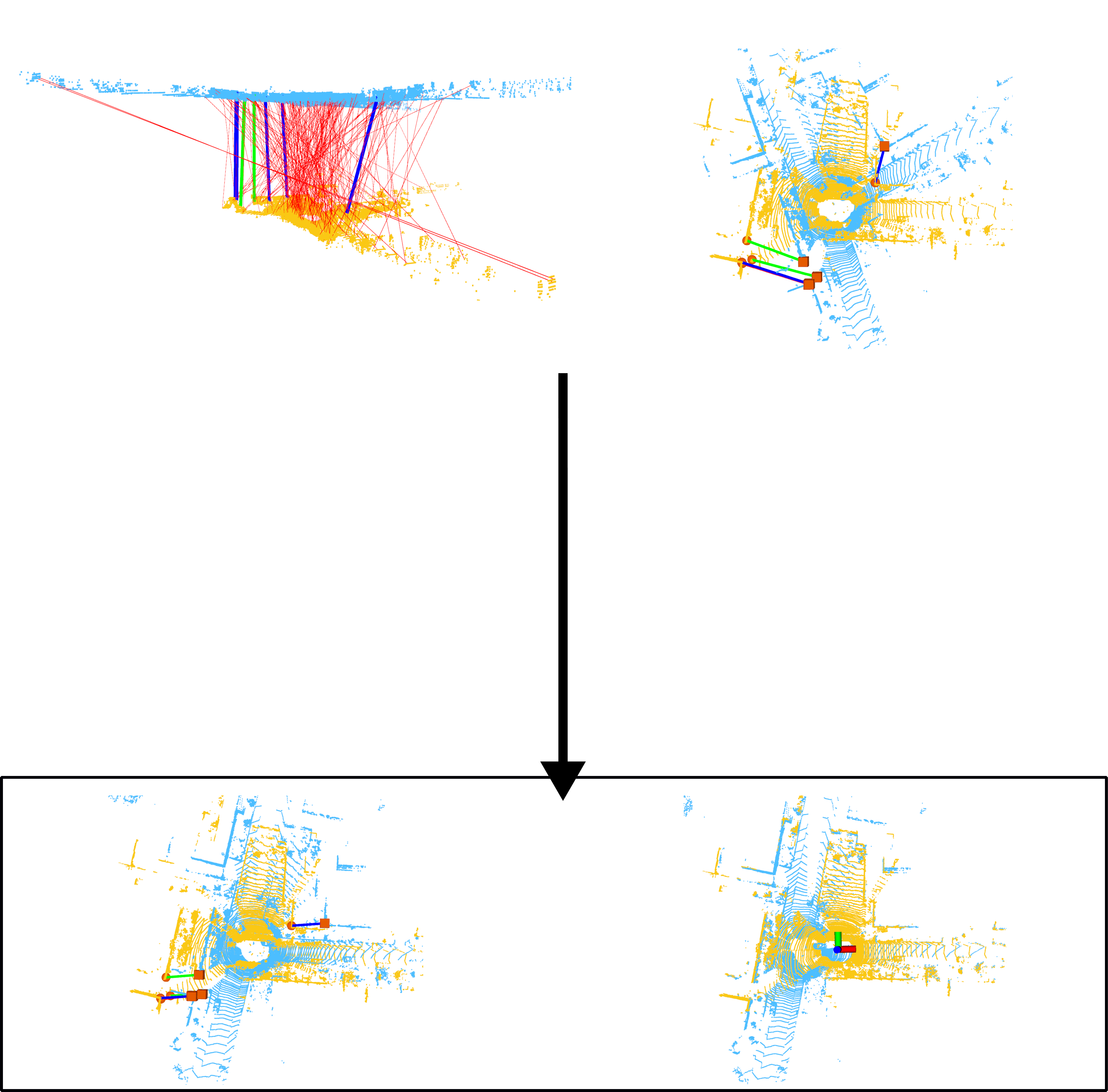
    \vspace{-0.05cm}
	\caption{Illustration of \textit{Quatro} in a degeneracy case when two distant and partially overlapped source (cyan) and target (yellow) are given. (a)~Spurious correspondences. (b)~The output of MCIS\textcolor{qwr}{-}\texttt{heuristic}. Most outliers are initially filtered. \textcolor{qwr}{(c)-(e)}~An example of Quasi-SO(3) estimation via GNC. \textcolor{qwr}{(c)}~First of all, all weights $w^{(0)}_k$ of TIMs are set to one. \textcolor{qwr}{(d)}~During the optimization, sometimes GNC unexpectedly leaves less than three pairs by assigning near-zero values to some $w^{(t)}_k$ (red dashed rectangle). \textcolor{qwr}{(e)} In the degeneracy case, quasi-SO(3) estimation is successfully done because DoF of $\mathbf{R}_{+}$ is one, so it can be estimated even when a single pair of TIMs is left. \textcolor{qwr}{(f)} Before and the after the application of COTE. For (a), (b), and \textcolor{qwr}{(f)}, the \textit{definite outliers}, inliers, and \textit{quasi-inliers} are represented by the red, green, and blue lines, respectively (best viewed in color).}
	\label{fig:illustration}
	\vspace{-0.6cm}
\end{figure}

\vspace{-0.15cm}

\noindent where $\tilde{r}_k$ denotes $r(\boldsymbol{\zeta}_k, \mathbf{R}_{+}^{(t)})$ for simplicity. The overall procedure is illustrated in Fig.~\ref{fig:illustration}. For each iteration, $\mu$ is updated as $\mu^{(t)} \leftarrow \kappa \cdot \mu^{(t-1)}$, where $\mu^{(0)}={\bar{c}^2}/({\max(r(\boldsymbol{\zeta}_k, \mathbf{I}_3))-\bar{c}^2})$ and \textcolor{qwr}{$\kappa > 1$} is a factor that increases the magnitude of non-convexity gradually, as presented in Fig.~\ref{fig:exp_evidences}(b). The iteration ends if the differential of $\sum_{k=1}^{K}\hat{w}_{k}^{(t)} r(\boldsymbol{\zeta}_k, \hat{\mathbf{R}}^{(t)}_{+})$ becomes small enough.

 
 \textcolor{comment}{Unfortunately, the outliers have a direct negative impact on the optimization at the first iteration because all the weights are initialized to~1, that is, $\hat{\mathbf{W}}^{(0)}=\mathbf{I}_K$ \textcolor{qwr}{(Fig.~\ref{fig:illustration}(c))}. Besides, in weight update, as in~(\ref{eq:truncated_weight}),  most weights are occasionally assigned to zero \textcolor{qwr}{(Fig.~\ref{fig:illustration}(d))}, resulting in degeneracy in the 3D space~\textcolor{qwr}{(Figs.~\ref{fig:illustration}(d) and ~\ref{fig:illustration}(g))}. However, $\mathbf{R}_{+}$ is designed to suppress the effect of outliers, as well as \textcolor{qwr}{to} be robust against degeneracy; thus, catastrophic failure can be prevented, as shown in \textcolor{qwr}{Fig.~\ref{fig:illustration}(e).}}
 
 
 \textcolor{comment}{In short, $\mathbf{R}_{+}$ has advantages over $\mathbf{R}$ for the following three reasons. First, estimation of $\mathbf{R}_{+}$ resolves the degeneracy issue in itself. As previously mentioned, the number of estimated inliers is not always guaranteed to be more than the minimum DoF of rotation in GNC. \textcolor{qw}{However, note that} the minimum DoF of $\mathbf{R}$ is three, whereas that of $\mathbf{R}_{+}$ is one. As a result, $\mathbf{R}_{+}$ can be more robust against degeneracy. Empirically, when the distance between the source and target viewpoints becomes \textcolor{comment}{more than 9 meters away}, \textcolor{qwr}{for example}, in outdoor environments, the degeneracy occasionally occurs. However, our method can conduct robust registration, even if the two viewpoints are farther apart (see Section~\rom{4}.\textit{A}).}
 
 \textcolor{comment}{Second, $\mathbf{R}_{+}$ helps to suppress the effect of the outliers. Because the ground can be consider\textcolor{qwr}{ed} locally flat enough and most structures tend to be orthogonal to the ground}, the local geometrical characteristics, e.g. surface normal, density, and so forth, are likely to be similar \textcolor{qwr}{along the normal} direction of the ground. Accordingly, the $\boldsymbol{\epsilon}_k$ can be decomposed into two terms: a) \textcolor{qwr}{a} parallel term to \textcolor{cv2}{the ground plane, i.e. $xy$-plane,} $\boldsymbol{\epsilon}^{||}_k$ and b) \textcolor{qwr}{a} perpendicular one $\boldsymbol{\epsilon}^{\perp}_k$, which satisfies $\boldsymbol{\epsilon}_k~=~\boldsymbol{\epsilon}^{||}_k~+~\boldsymbol{\epsilon}^{\perp}_k$. Consequently, the outliers are classfied into two groups. One is the \textit{quasi-inliers}, $\mathcal{Q}$, which satisfies {$\abs{\boldsymbol{\epsilon}^{||}_k} \approx 0$ and \textcolor{qwe}{$\abs{\boldsymbol{\epsilon}^{\perp}_k} \gg 0$}}, and the other is the \textit{definite outliers}, $\mathcal{D}$, which satisfies \textcolor{qwe}{$ \abs{\boldsymbol{\epsilon}^{||}_k} \gg 0$}. Note that $\hat{\mathcal{Q}} \cup \hat{\mathcal{D}} = {\hat{\mathcal{O}}}$, where $\hat{}$ denotes the \textcolor{cv2}{estimated output of GNC}. In that context, $\boldsymbol{\epsilon}^{\perp}_k$ of quasi-inliers rarely affect the estimation of $\mathbf{R}_{+}$ because the estimation of $\mathbf{R}_{+}$ is invariant to $z$-values. Thus, $\mathcal{Q}$ can be considered as additional inliers when optimizing (\ref{eq:black_rangarajan}).


Third, as an extension of the second reason, $\mathcal{Q}$ has a positive impact on divergence prevention by increasing the ratio of inliers (i.e. $\frac{|{\hat{\mathcal{O}}^C}|}{|\hat{\mathcal{O}}|}$ to $\frac{|{\hat{\mathcal{O}}^C} \cup \hat{\mathcal{Q}}|}{|\hat{\mathcal{D}}|}$). This is because GNC-based optimization sometimes diverge when the ratio of the outlier correspondences is too large \cite{yang2020gnc,shi2020robin}. 


\subsection{Component-wise Translation Estimation}

Finally, the relative translation is estimated in a component-wise way, as shown in \textcolor{qwr}{Fig.~\ref{fig:illustration}(f)}. Let the boundary interval set ${^{l}\mathcal{E}}$ be the $2\abs{\mathcal{A}}$-tuples that comprises the lower bound, ${^{l}\mathbf{v}}_{ij} - \sigma_{ij}\bar{c}$, and the upper bound, $^{l}\mathbf{v}_{ij} + \sigma_{ij}\bar{c}$, and assume that all the elements of $^{l}\mathcal{E}$ is sorted in ascending order. Then, let the $g$-th consensus set be $^{l}\mathcal{I}_g=\{(i,j)| \frac{({^{l}\phi_{g}}-{^{l}\mathbf{v}}_{ij})^{2}}{\sigma_{ij}^{2}} \leq \bar{c}^{2}\}$, where ${^{l}\phi_{g}}\in\mathbb{R}$ is any value that satisfies \textcolor{qwr}{${^{l}\mathcal{E}}(g)<{^{l}\phi_{g}}<{^{l}\mathcal{E}}(g+1)$} for $g=1,2,\dots,2\abs{\mathcal{A}}-1$. Then, $^{l}{\hat{\mathbf{t}}}_g$ is estimated by the weighted average for non-empty $^{l}\mathcal{I}_g$ as follows:

\vspace{-0.25cm}

\begin{equation}
    {^{l}{\hat{\mathbf{t}}}_{g}}=\Big(\sum_{(i,j) \in ^{l}\mathcal{I}_{g}} \frac{1}{\sigma_{ij}^{2}}\Big)^{-1}\sum_{(i,j) \in ^{l}\mathcal{I}_{g}} \frac{^{l}\mathbf{v}_{ij}}{\sigma_{ij}^{2}}.
\end{equation}

\vspace{-0.15cm}

\noindent Finally, among up to $2\abs{\mathcal{A}}-1$ candidates~$^{l}\hat{\mathbf{t}}_g$, \textcolor{qwr}{${^l\hat{\mathbf{t}}}$ is selected} which minimizes the truncated objective function~(\ref{eqn:decouple_trans}). Note that COTE is based on the premise that the estimated rotation is precise enough. 


Therefore, our method finally enables degeneracy-robust global registration even though a single pair is left due to poor feature matching and drastic correspondence pruning.

\subsection{Preprosessing of Correspondences}

 In this paper, \textcolor{qwr}{fast point feature histogram~(FPFH)}~\cite{rusu2009fpfh} is utilized, \textcolor{cv2}{which is widely used as a conventional descriptor for the registration~\cite{zhou2016fast}}. However, the original FPFH for a 3D point cloud captured by a 64-channel LiDAR sensor takes tens of seconds, which is too slow. For this reason, we employ voxel-sampled FPFH, which is preceded by voxel-sampling with voxel size~$\nu$. This is followed by the correspondence test~\cite{zhou2016fast} that outputs $\mathcal{A}_{\text{raw}}$, as represented in Fig.~\ref{fig:illustration}(a). 

Thereafter, the \textcolor{qwr}{max clique inlier selection}~(MCIS)~\cite{rossi2013pmc} is applied, which takes $\mathcal{A}_{\text{raw}}$ as input and outputs $\mathcal{A}$, as shown in Fig.~\ref{fig:illustration}(b). TEASER++~\cite{yang2020teaser} is the representative method to adopt the MCIS first, and the researchers showed that MCIS successfully discards gross outliers. We found that finding the exact maximal clique is the bottleneck of the algorithm in terms of speed. Thus, we revise it to find a heuristic maximal clique whose cardinality is large enough, which is \textcolor{qwr}{named} as MCIS-\texttt{heuristic}.

\begin{figure*}[t!]
    \captionsetup{font=footnotesize}
	\centering
	\begin{subfigure}[b]{0.18\textwidth}
		\includegraphics[width=1.0\textwidth]{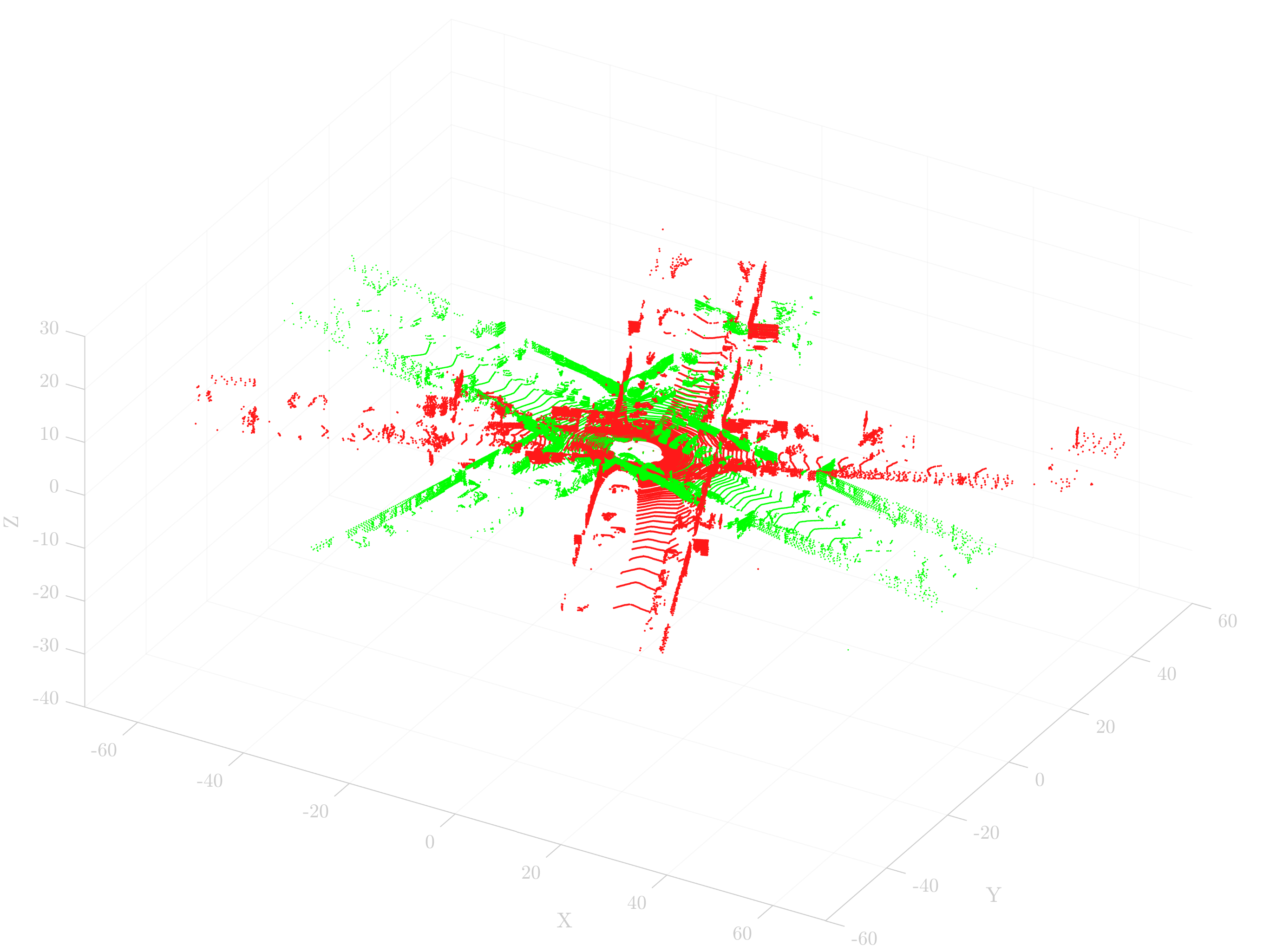}
		\includegraphics[width=1.0\textwidth]{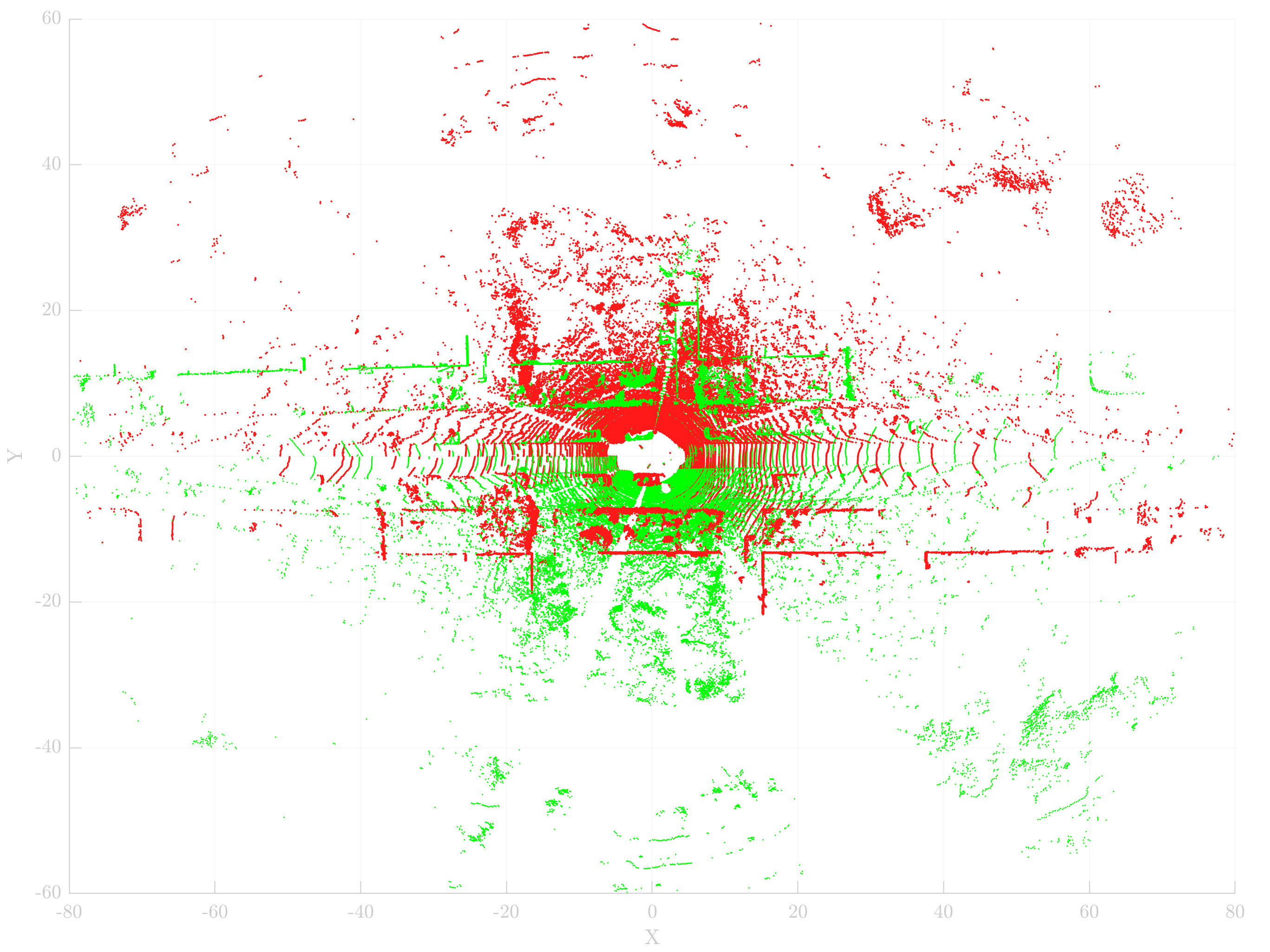}
		\caption{Source and target}
	\end{subfigure}
	\begin{subfigure}[b]{0.18\textwidth}
		\includegraphics[width=1.0\textwidth]{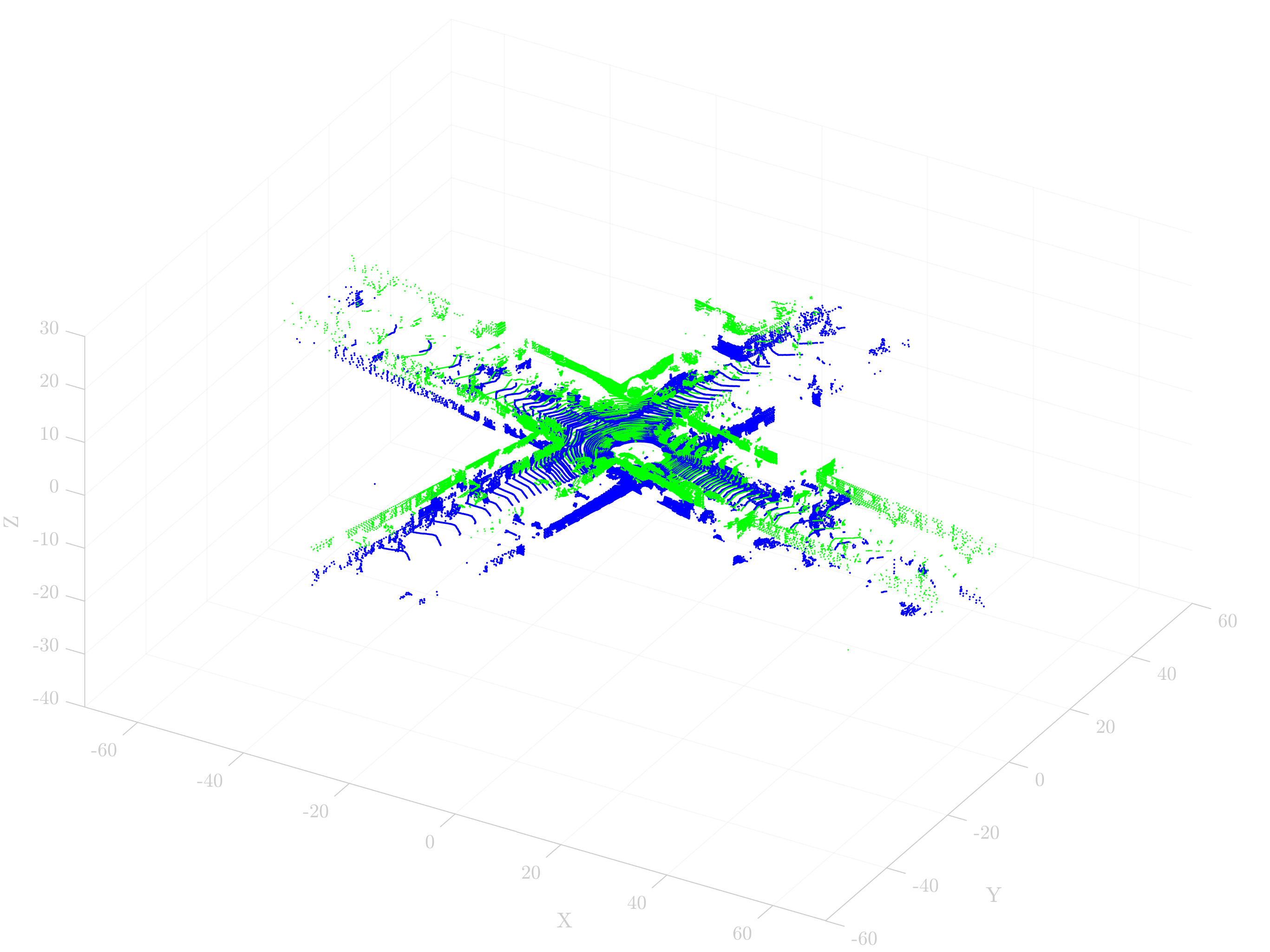}
		\includegraphics[width=1.0\textwidth]{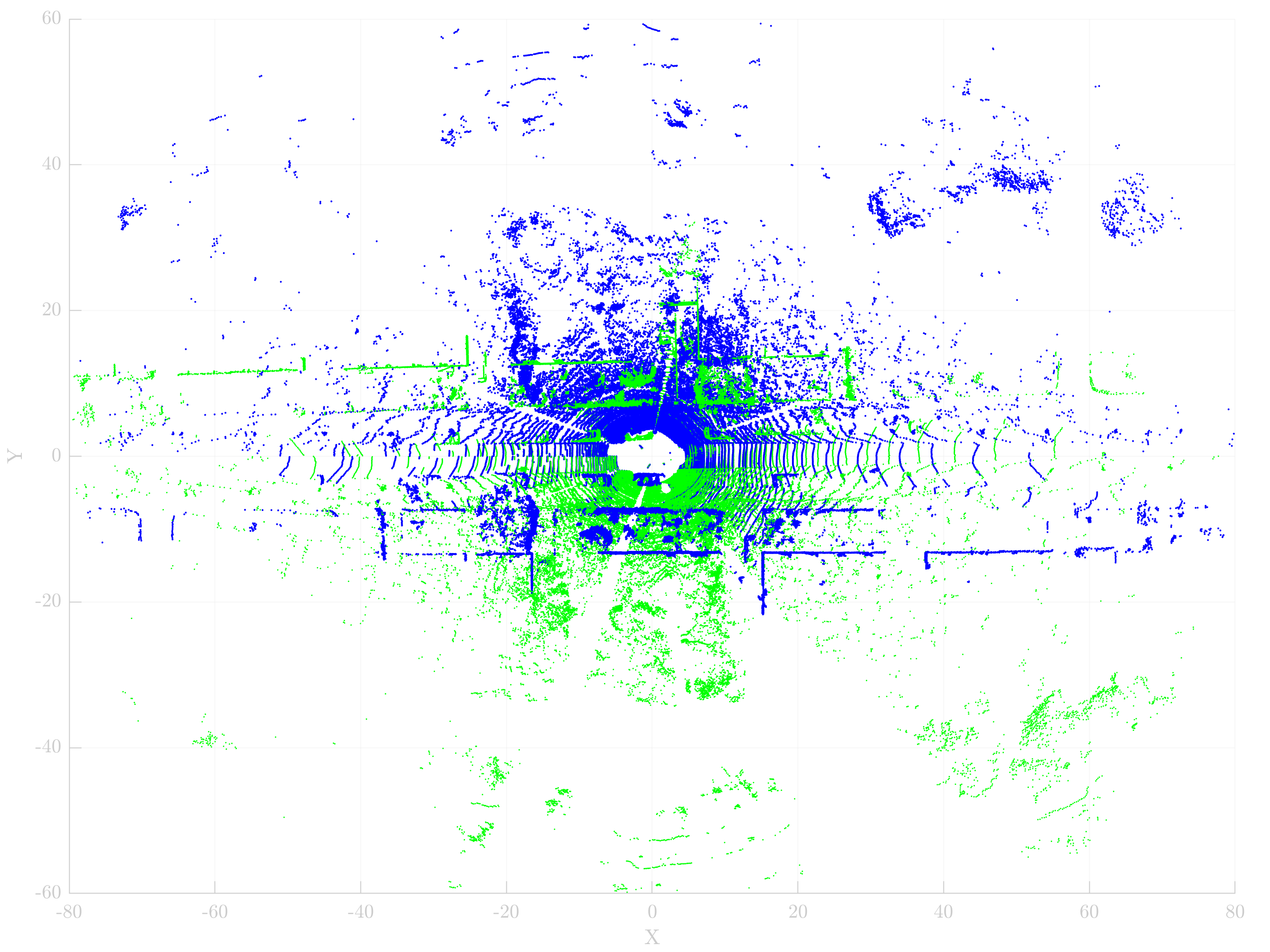}
		\caption{RANSAC10K~\cite{fischler1981ransac}}
	\end{subfigure}
	\begin{subfigure}[b]{0.18\textwidth}
		\includegraphics[width=1.0\textwidth]{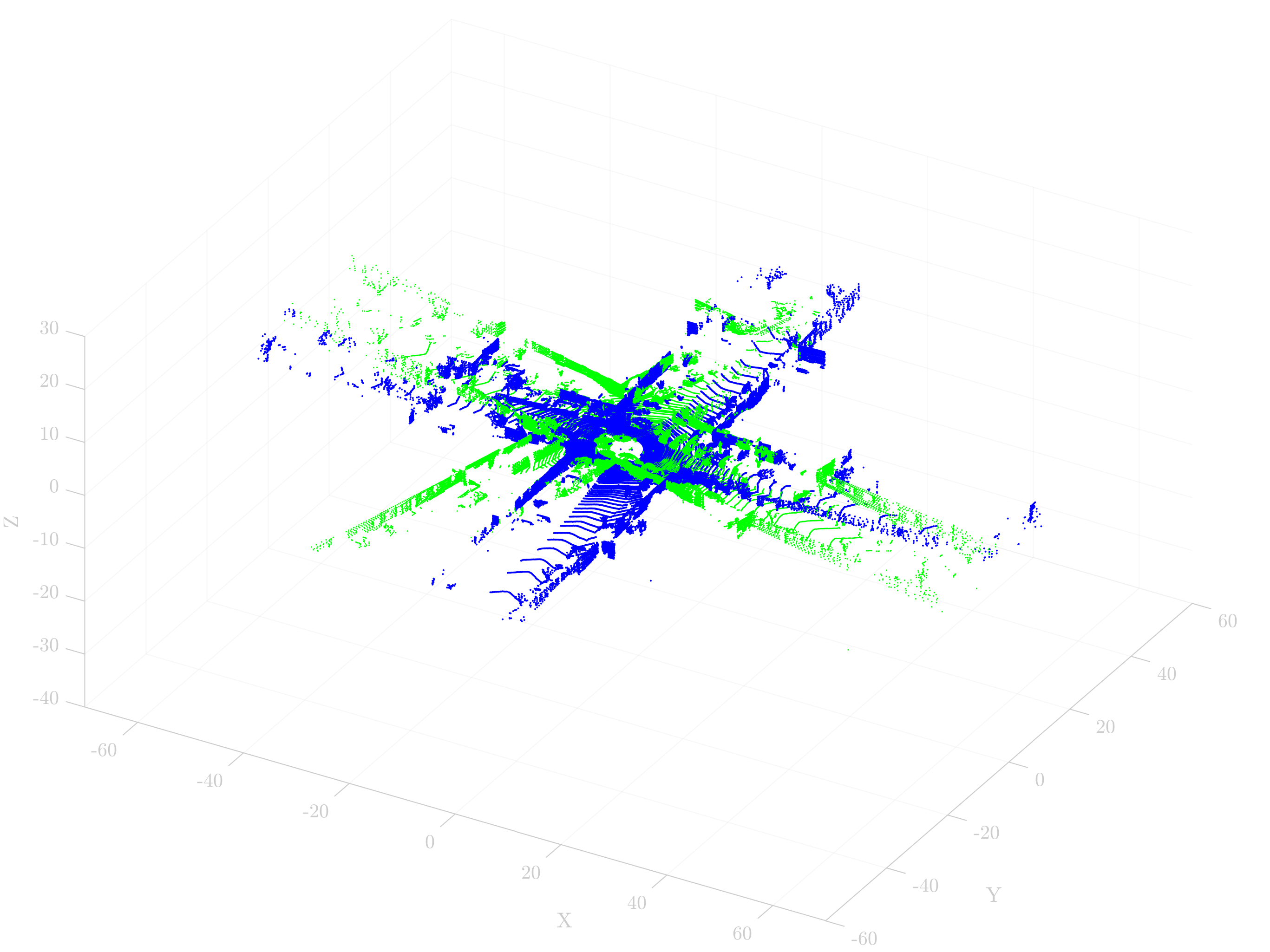}
		\includegraphics[width=1.0\textwidth]{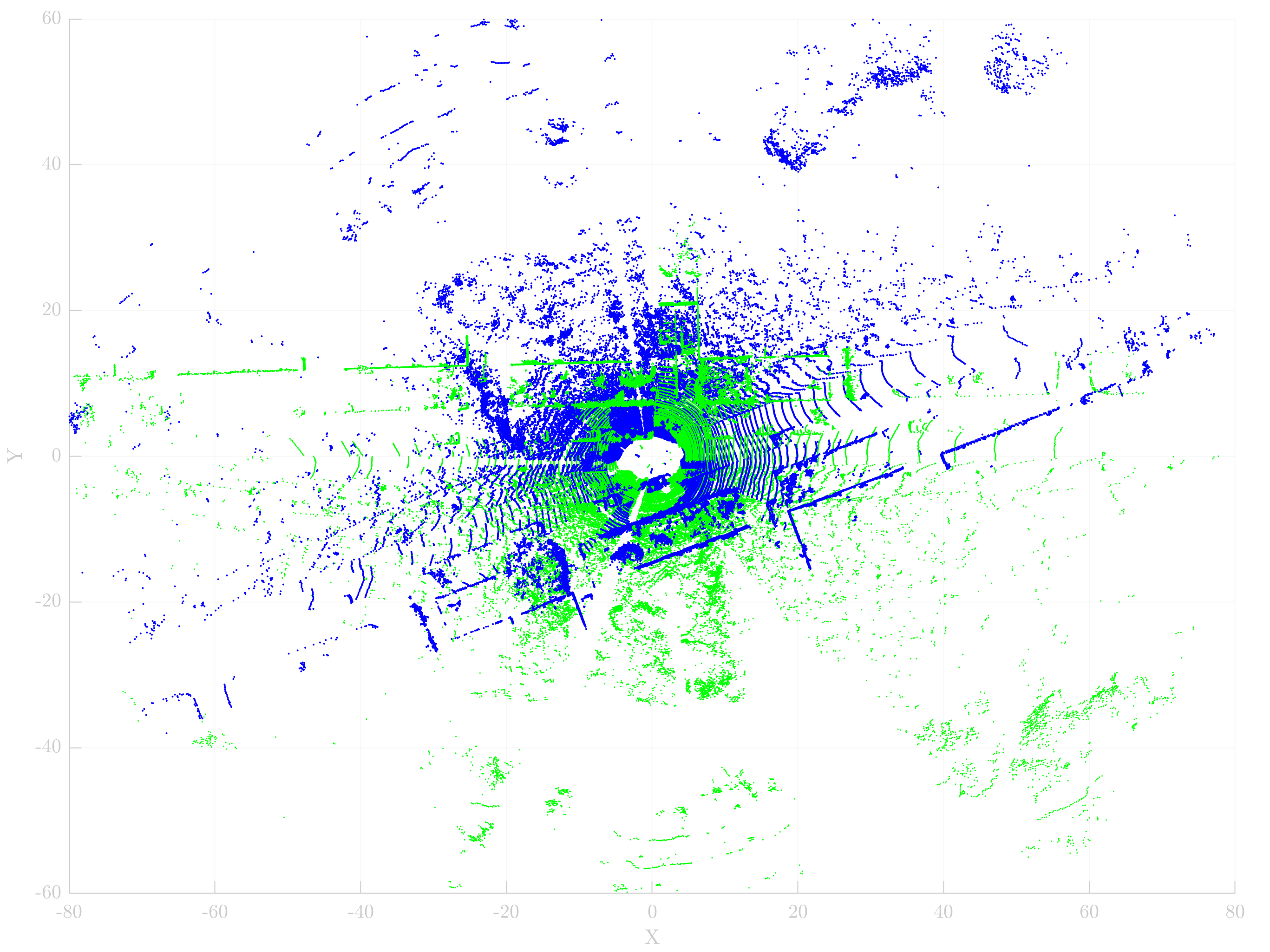}
		\caption{FGR \cite{zhou2016fast}}
	\end{subfigure}
	\begin{subfigure}[b]{0.18\textwidth}
		\includegraphics[width=1.0\textwidth]{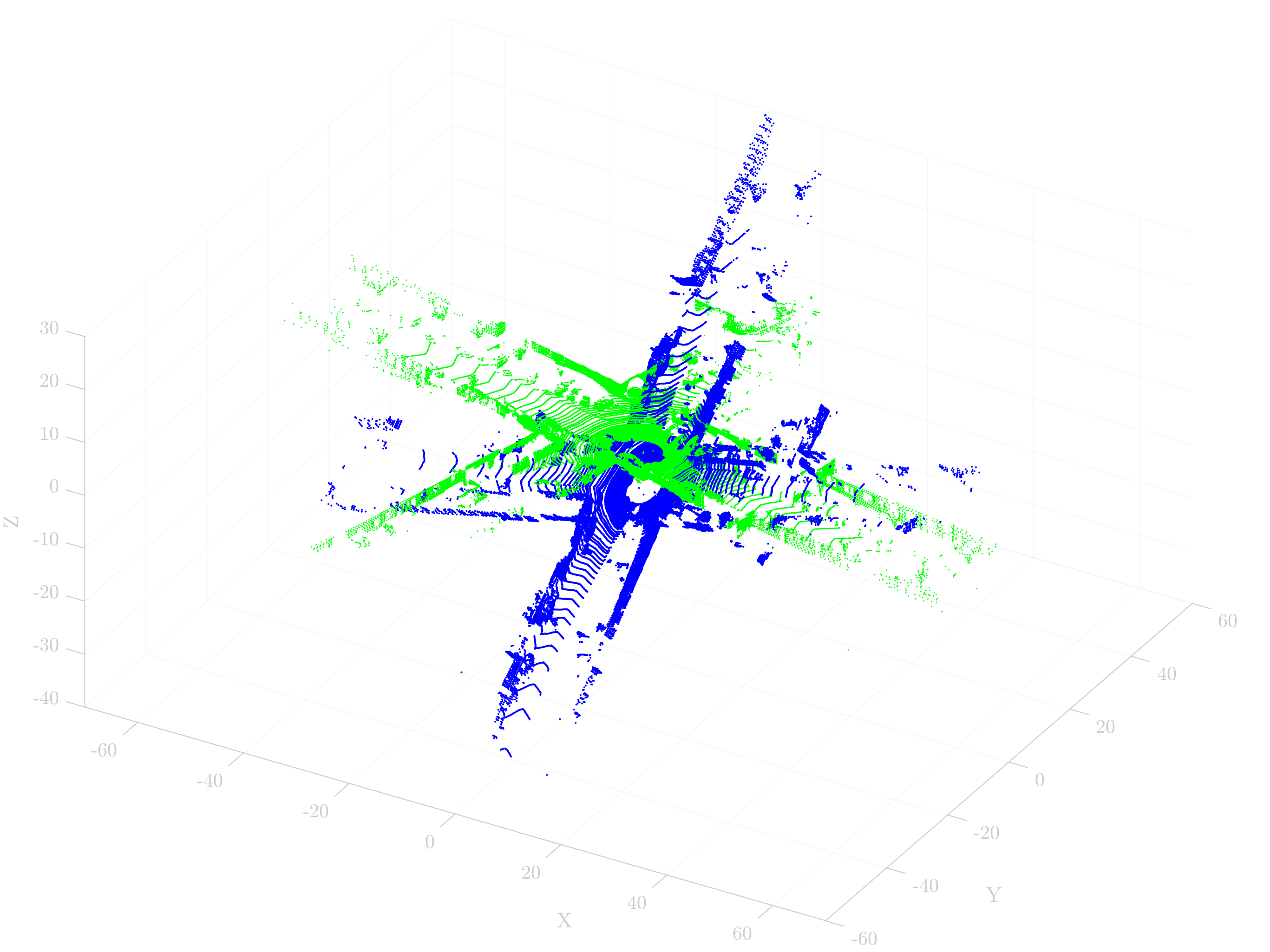}
		\includegraphics[width=1.0\textwidth]{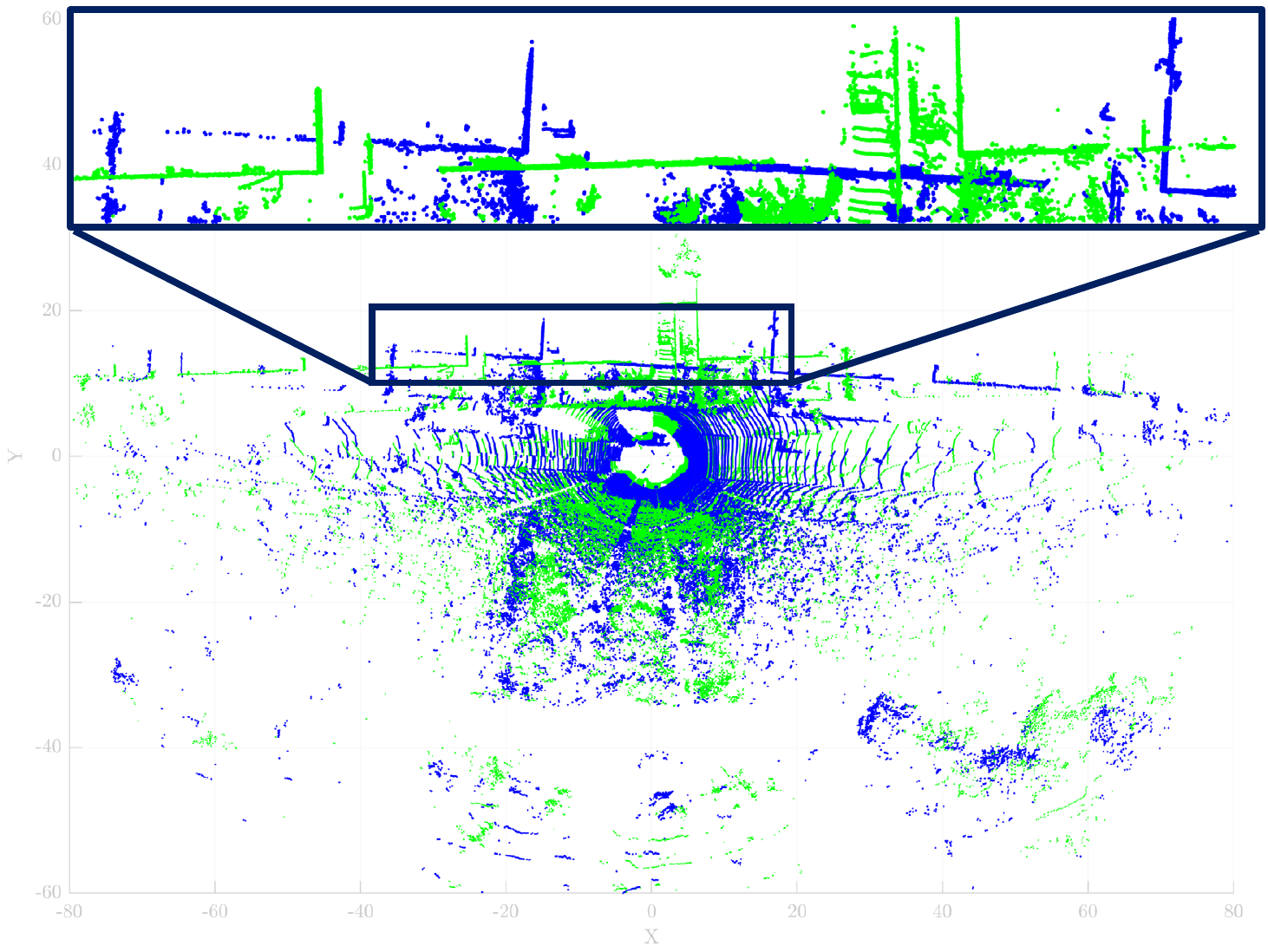}
		\caption{TEASER++ \cite{yang2020teaser}}
	\end{subfigure}
	\begin{subfigure}[b]{0.18\textwidth}
		\includegraphics[width=1.0\textwidth]{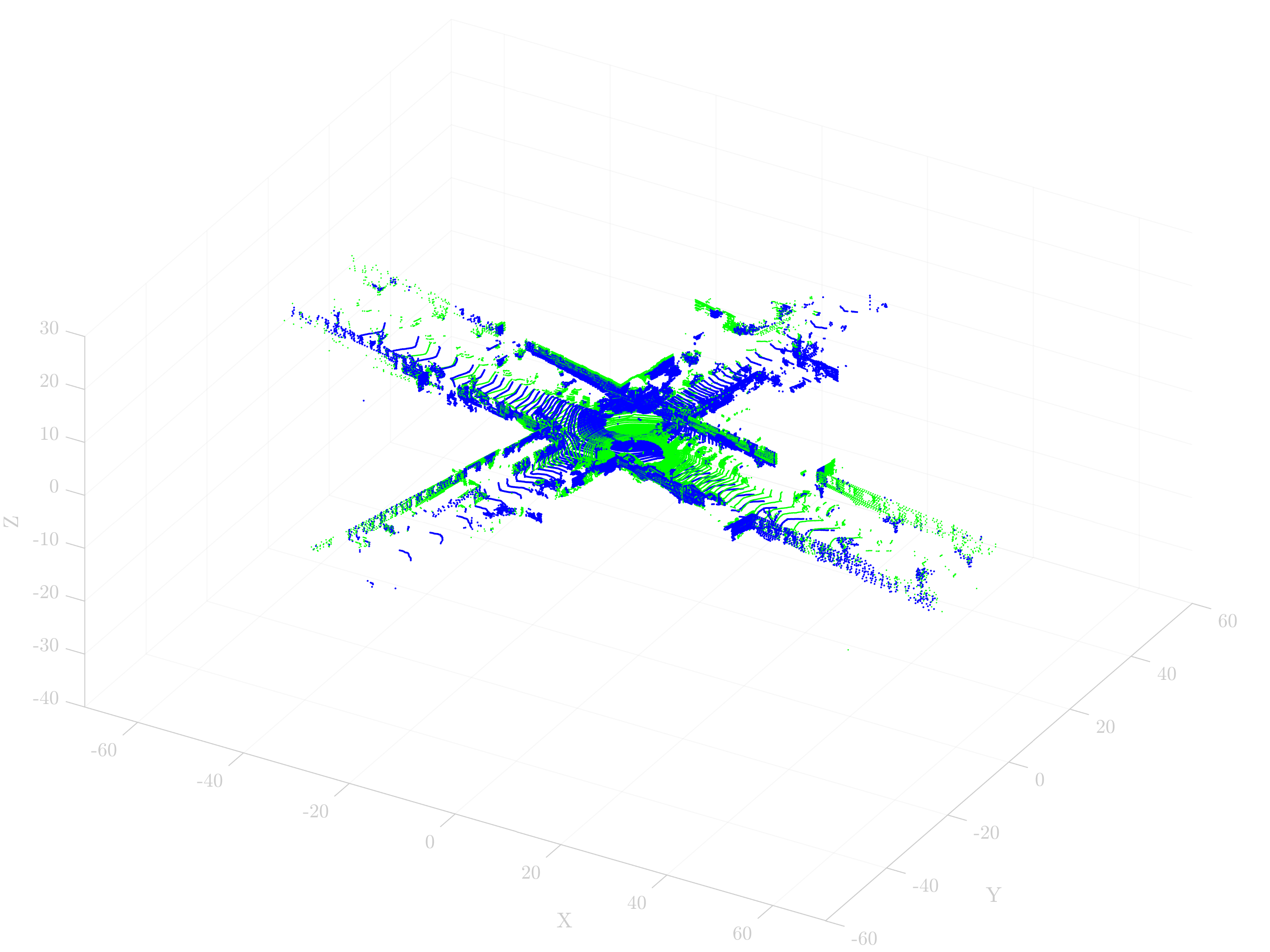}
		\includegraphics[width=1.0\textwidth]{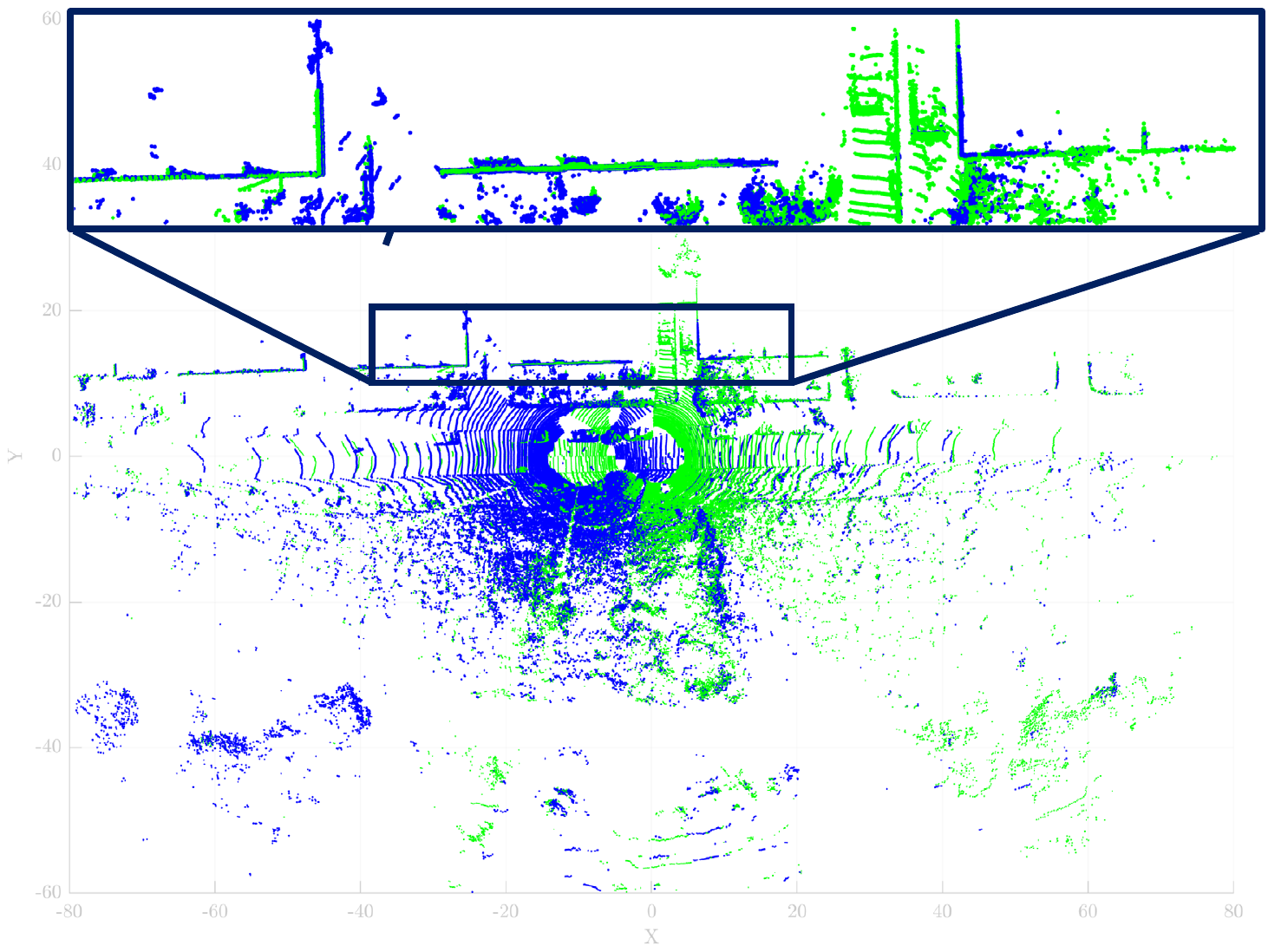}
		\caption{Quatro (Ours)}
	\end{subfigure}
	\vspace{-0.2cm}
	\caption{Degeneracy cases (T-B): Registration results between 1,398 to 3,554 frames in Seq.~\texttt{00}, which are 9.88 m apart, and between 177 and 1,726 frames in Seq.~\texttt{08}, which are 9.82 m apart. The red, green, and blue points denote source, target, and estimate, respectively. Note that RANSAC~\cite{fischler1981ransac} returns $\mathbf{I}_4$ if it fails to find correct putative correspondences. The closer the green and blue points are, the better (best viewed in color).}
	\label{fig:kitti_qualitative}
	\vspace{-0.7cm}
\end{figure*}


\section{Experiments}
\vspace{-0.1cm}
\subsection{Dataset}

We experimented in the indoor/outdoor environments, i.e. KITTI dataset \cite{geiger2012kitticvpr,geiger2013vision} and \textcolor{qw}{NAVER LABS localization dataset} \cite{lee2021large_naverlabs}. In particular, we tested the algorithms in a)~loop-closing situations, b)~augmented rotation situations, and c)~odometry situations.

\subsection{Error Metrics}

As quantitative metrics, the average relative pose errors, $t_{\text{avg}}$ \textcolor{qwr}{for translation} and $r_{\text{avg}}$ \textcolor{qwr}{for rotation}, are used as follows:

\begin{itemize}
    \item $t_{\text{avg}}= \sum_{n=1}^{N} (\mathbf{t}_{n, \text{GT}}-{\hat{\mathbf{t}}}_{n})^{2} / N$,
    \item $r_{\text{avg}}= \frac{180}{\pi} \cdot \sum_{n=1}^{N} \cos^{-1} (\frac{\operatorname{Tr}\left({\hat{\mathbf{R}}}_{n}^{\intercal} \mathbf{R}_{n, \text{GT}}\right)-1}{2}) / N  $ 
\end{itemize}
\noindent where the subscript GT denotes that the value is from the ground truth. For the odometry test, the relative odometry errors,  $t_\text{rel}$ and $r_\text{rel}$, are used, which are calculated by \cite{Zhang18rpg_eval_tool}.

\subsection{Parameters of Quatro} 

Empirically, we set $\bar{c}= 0.15$ and $\kappa=1.4$. Because of the sparsity differences depend on the number of channels of LiDAR sensors, the parameters for voxel-sampled FPFH should be changed depending on the sensor configuration. Thus, we set $\nu = 0.3~\text{m}$, radius for normal estimation $r_{\text{normal}}=0.5~\text{m}$, and radius for FPFH $r_{\text{FPFH}}=0.65~\text{m}$ in the KITTI dataset~\cite{geiger2012kitticvpr}, and $\nu = 0.1~\text{m}$, $r_{\text{normal}}=0.3~\text{m}$, and $r_{\text{FPFH}} = 0.45~\text{m}$ in the \textcolor{qw}{NAVER LABS localization} dataset~\cite{lee2021large_naverlabs}.
\vspace{-0.3cm}
\section{Experimental Results}
\vspace{-0.15cm}

 To check the effectiveness of our proposed method, Quatro was quantitatively compared with state-of-the-art methods, namely, RANSAC~\cite{fischler1981ransac}, (specifically, RANSAC1K and RANSAC10K, \textcolor{qwr}{where} each number denotes the number of iteration), FGR~\cite{zhou2016fast}, and TEASER++~\cite{yang2020teaser}. We leveraged the open-source implementations for the experiment. \textcolor{comment}{Quatro-\texttt{c2f} comprises the proposed Quatro as a coarse alignment, followed by G-ICP~\cite{koide2020vgicp} as a fine alignment.}


\vspace{-0.1cm}
\subsection{Impact of Quasi-SE(3) in Distant Global Registration}
\vspace{-0.1cm}

\begin{figure}[t!]
    \captionsetup{font=footnotesize}
	\centering 
	\begin{subfigure}[b]{0.46\textwidth}
		\includegraphics[width=1.0\textwidth]{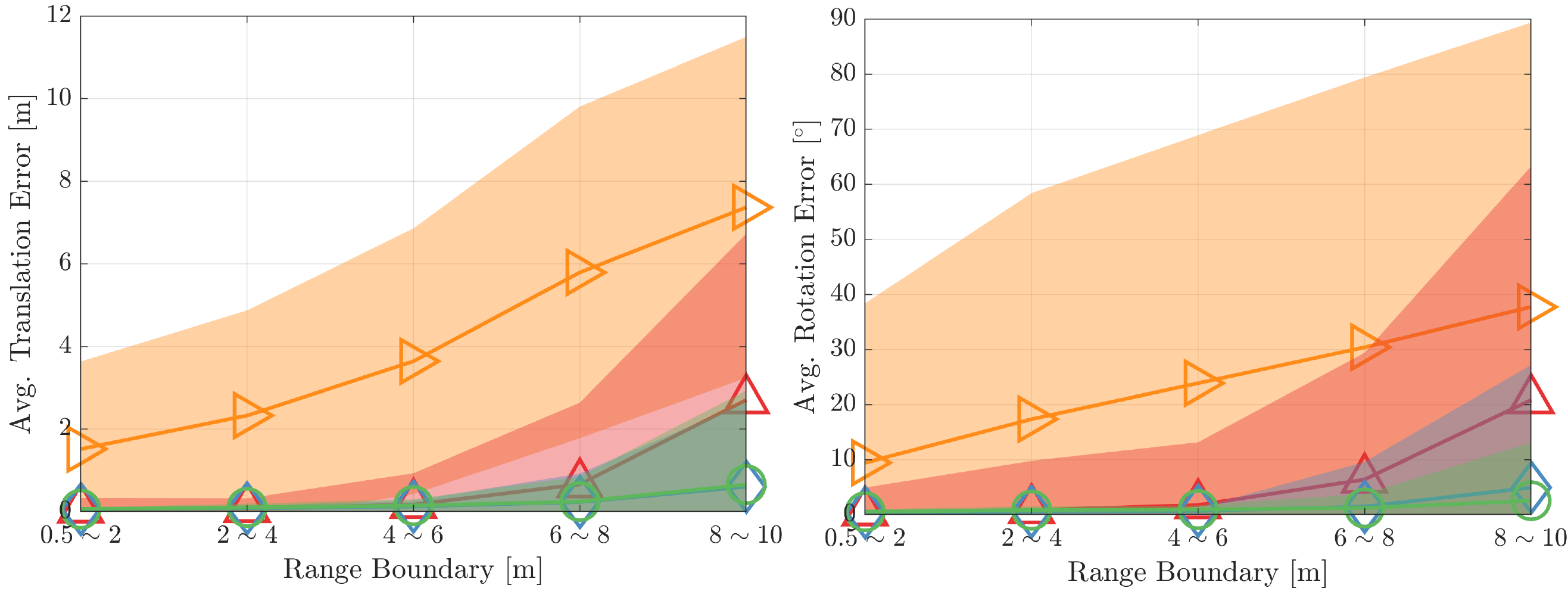}
		\vspace{-0.7cm}
		\caption{}
	\end{subfigure}
	\begin{subfigure}[b]{0.46\textwidth}
		\includegraphics[width=1.0\textwidth]{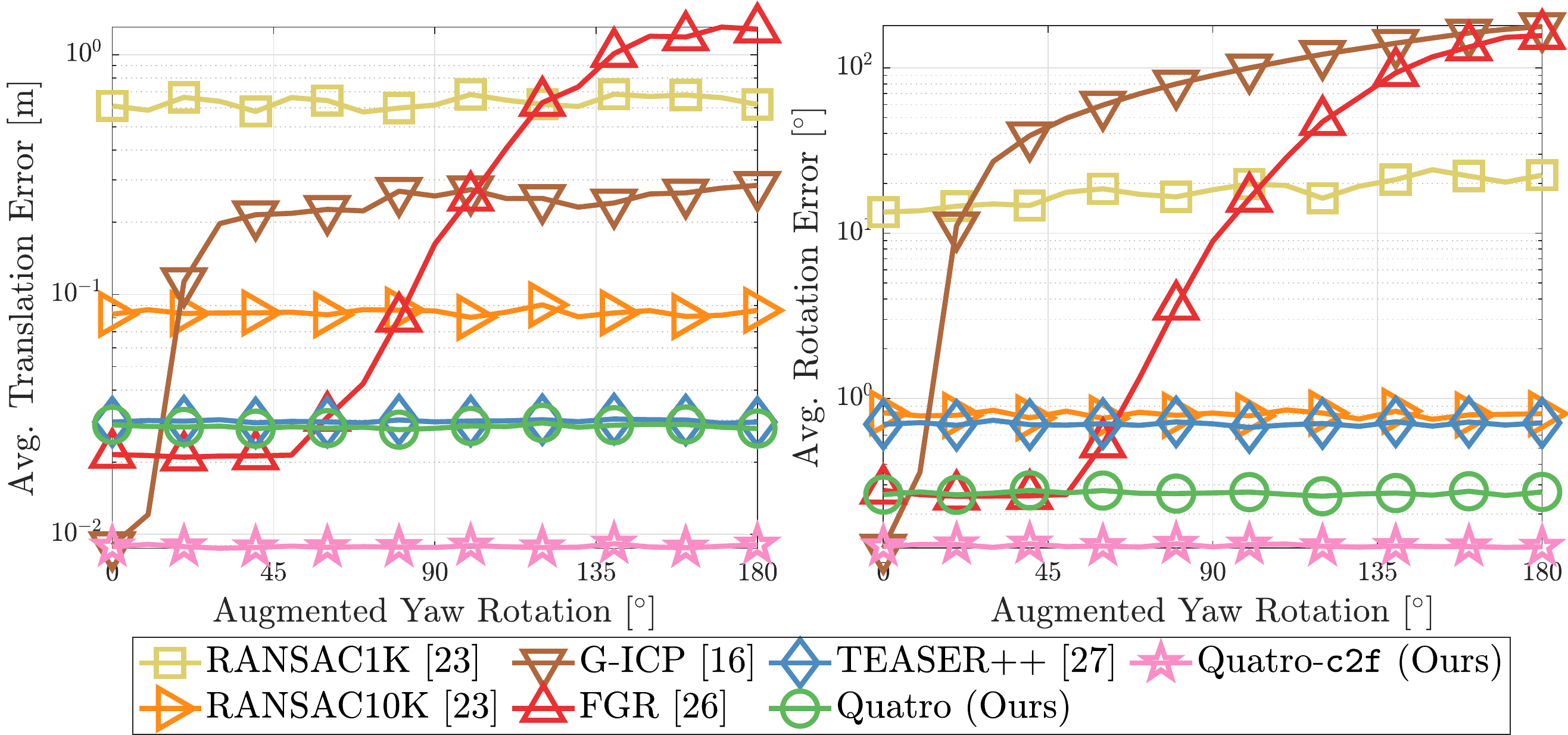}
		\vspace{-0.5cm}
		\caption{}
	\end{subfigure}
	\vspace{-0.3cm}
	\caption{Performance changes with respect to (a) the range boundary in loop-closing situations in the KITTI dataset~\cite{geiger2012kitticvpr} and (b) the augmented yaw rotation when the frame interval is set to 5 (i.e. $\Delta 5$) in \texttt{HD\_B1} of the \textcolor{qw}{NAVER LABS localization} dataset~\cite{lee2021large_naverlabs}.}
	\label{fig:ablation}
	\vspace{-0.6cm}
\end{figure}

In general, the state-of-the-art methods estimate precise relative pose by overcoming the effect of gross outliers. However, our Quatro exhibits noticeable robustness in the case where the relative pose between two viewpoints of source and target is distant, as shown in Fig.~\ref{fig:kitti_qualitative} and Table~\ref{table:success_rate}. In particular, it was observed that FGR was more likely to fail to \textcolor{qw}{conduct} registration compared with TEASER++ and our Quatro. This is because FGR inherently uses linearization of SE(3) in optimization. Thus, if the linearization assumption does not hold, its performance becomes degraded, as shown in Fig.~\ref{fig:ablation}. Furthermore, FGR has no additional outlier rejection procedure, i.e. MCIS. Consequently, as the ratio of outliers increases, the performance obviously decreases.

\begingroup
\begin{table}[t!]
	\centering
	\captionsetup{font=footnotesize}
	\caption{Success rate \textcolor{qwr}{(unit: \%) for situations where distance is between 9$\sim$12 m away under the loop-closing condition} in the KITTI dataset~\cite{geiger2012kitticvpr}. The registration result was considered successful when the transition
	\textcolor{qwr}{and rotation errors were not larger than 2~m and 10$^\circ$, respectively}. The criteria \textcolor{qwr}{were set based on~\cite{kim2019gp}}: a pose difference sufficient for the local registration to allow the estimate to converge toward global minima.}
	\vspace{-0.15cm}
	\setlength{\tabcolsep}{5pt}
	\begin{tabular}{l|cccccc}
	
	\toprule \midrule
	Method & \texttt{00} & \texttt{02} & \texttt{05} & \texttt{06} & \texttt{07} &  \texttt{08}  \\ \midrule
	FGR~\cite{zhou2016fast} & 43.81 & 31.45 & 51.37 & 56.46 & 44.49 & 15.78  \\
 	TEASER++~\cite{yang2020teaser} & 98.62 & 98.87 & 98.67 & 97.93 & 97.90 & 98.63  \\
 	Quatro (Ours) & \textbf{99.30} & \textbf{99.24} & \textbf{99.63} & \textbf{99.34} & \textbf{99.75} & \textbf{98.86}  \\ \midrule\bottomrule
	\end{tabular}
	\label{table:success_rate}
	\vspace{-0.3cm}
\end{table}
\endgroup

On the other hand, TEASER++, which employs MCIS similar to our proposed method, shows robust performance compared with FGR in the distant cases. Unfortunately, MCIS also occasionally rejected numerous matching pairs, so it was no longer guaranteed that there are more than three estimated inliers left in the estimating rotation. Note that TEASER++ also utilized decoupling of rotation and translation, so if TEASER++ fails to estimate the rotation, then it inevitably fails to estimate translation as a domino effect, as mentioned in Section~\rom{2}.\textit{E}. Thus, TEASER++ shows relatively large variance as the range boundary becomes larger, compared with our Quatro, as shown in \textcolor{qwr}{Fig.~\ref{fig:ablation}(a)}. 

In particular, the variance difference between TEASER++ and Quatro in estimating rotation becomes significant, as presented in \textcolor{qwr}{Fig.~\ref{fig:ablation}(a)}. This is because our quasi-SE(3) reduces its minimum \textcolor{qw}{number of the required pairs} to one via quasi-SO(3) estimation and COTE. Accordingly, it successfully avoids degeneracy, even if \textcolor{qwr}{it is} given less than three estimated inliers. As a result, catastrophic failure in the rotation estimation can be somewhat prevented. Furthermore, \textcolor{qwr}{due to} the effect of the quasi-inliers, \textcolor{qw}{some irregular errors} (i.e. $\epsilon_{k}^{\perp}$) no longer affect the estimation of $\mathbf{R}_{+}$.Thus, our method generally has a success rate of more than 99\%. 

Therefore, these experimental results corroborate that our quasi-SE(3) estimation enables robust global registration, while overcoming the degeneracy problem.
 
\subsection{Comparison in Loop Closing/Odometry Test}

As explained eariler, some registration methods show performance degradation as the distance or difference of rotation between two viewpoints of source and target become\textcolor{qwr}{s} larger, as shown in Fig.~\ref{fig:ablation}. In contrast, our proposed method shows promising performance, especially with little variance in performance relative to other methods. Furthermore, it was shown that our quasi-SE(3) is additionally advantageous in indoor situations where the ground is mostly flat and the urban structures are orthogonal to the ground; thus, our assumption $\angle \mathbf{R}_y \cdot \mathbf{R}_x \approx \mathbf{I}_{3}$ is obviously met. As a result, our method maintains the smallest rotation error, even though source and target clouds are in diametrically opposite directions.
 
 Furthermore, our method shows better odometry results, as shown in Table~\ref{table:kitti_comparison}. It is natural that local registration methods show better performance when the frame interval is small, but as the interval becomes larger, their performance drastically degraded. In contrast, the global registration methods show robust performance even though the frame interval becomes larger. In particular, performance of our method decreases little, showing smaller odometry errors than the other global registration methods.

\subsection{Quatro as a Coarse Alignment}
 
Finally, Quatro-\texttt{c2f} shows successful coarse-to-fine registration, as shown in Fig.~\ref{fig:ablation}(b) and Table~\ref{table:kitti_comparison}. In particular, Quatro-\texttt{c2f} even shows better performance compared with the state-of-the-art methods, including conventional and deep learning-based approaches in a sequence that is used as \textcolor{qwr}{a} training dataset. Therefore, we finally confirm that Quatro can provide a sufficiently accurate coarse alignment, thus helping local registration algorithms conduct fine alignment successfully.

\begingroup
\begin{table}[t!]
    \captionsetup{font=footnotesize}
	\centering
	\caption{Comparison of odometry test with the state-of-the-art methods on Seq.~\texttt{00} of the KITTI dataset~\cite{geiger2012kitticvpr}. The results of deep learning-based methods are from \cite{fischer2021stickypillars,li2020dmlo} ($t_\text{rel}$:~\%, $r_\text{rel}$:~$\deg$/100m).}  
	\vspace{-0.15cm}
	\setlength{\tabcolsep}{4pt}
	
	{\scriptsize
	
	\begin{tabular}{l|l|cccccc}
	
	\toprule \midrule
	&\multirow{2}[3]{*}{Method} & \multicolumn{2}{c}{$\Delta 1$} & \multicolumn{2}{c}{$\Delta 3$} & \multicolumn{2}{c}{$\Delta 5$} \\  \cmidrule(lr){3-4} \cmidrule(lr){5-6} \cmidrule(lr){7-8} 
	&  & $t_{\text{rel}}$ & $r_{\text{rel}}$ & $t_{\text{rel}}$ & $r_{\text{rel}}$ & $t_{\text{rel}}$  & $r_{\text{rel}}$   \\ \midrule
	\parbox[t]{2mm}{\multirow{3}{*}{\rotatebox[origin=c]{90}{Local}}} &ICP~\cite{besl1992method} & 6.88 & 2.99 & 21.92 & 8.70 & 21.14 & 8.51 \\
	&G-ICP~\cite{segal2009gicp} & 1.26 & 0.45 & 5.50 & 1.45 & 14.20 & 3.32\\
	&VG-ICP~\cite{koide2020vgicp} & \textbf{1.03} & \textbf{0.30} & 11.83 & 1.65 & 19.11 & 6.32  \\ \midrule
	\parbox[t]{2mm}{\multirow{3}{*}{\rotatebox[origin=c]{90}{Global}}}&
	FGR~\cite{zhou2016fast}& 2.73 & 0.69 & 7.17 & 1.58 & 14.66 & 4.12  \\
	&TEASER++~\cite{yang2020teaser} & 2.11 & 0.91 & 2.64 & 1.11  & 3.19 & 0.91 \\  
	&Quatro (Ours) & 1.45 & 0.41 & \textbf{1.38} & \textbf{0.24} & \textbf{1.94} & \textbf{0.46}  \\ \midrule \midrule
	\parbox[t]{2mm}{\multirow{5}{*}{\rotatebox[origin=c]{90}{Deep}}}&LO-Net~\cite{li2019net} & 1.47 & 0.72 & N/A & N/A & N/A & N/A  \\
	&LO-Net+\texttt{M}~\cite{li2019net} & 0.78 & 0.42 & N/A & N/A & N/A & N/A  \\
	&DMLO$^\dagger$~\cite{li2020dmlo} & 0.83 & 0.44 & N/A & N/A & N/A & N/A  \\
	&DMLO+\texttt{M}$^\dagger$~\cite{li2020dmlo} & 0.73 & 0.44 & N/A & N/A & N/A & N/A  \\
	&StickyPillars$^{\dagger, \mathsection}$~\cite{fischer2021stickypillars} & \textbf{0.65} & 0.26 & 0.79 & 0.31 & 1.29 & 0.48  \\\midrule
	\parbox[t]{2mm}{\multirow{3}{*}{\rotatebox[origin=c]{90}{Conv.}}}  &SUMA~\cite{behley2018efficient} & 0.68& 0.23 & 1.69 & 0.61 & 2.36 & 0.51 \\ 
	&A-LOAM~\cite{zhang2014loam} & 0.70 & 0.27 & 0.97 & 0.38 & 31.16 & 12.10  \\
	&Quatro-\texttt{c2f} (Ours) & \textbf{0.65} & \textbf{0.21} & \textbf{0.67} & \textbf{0.21} & \textbf{0.67} & \textbf{0.21}  \\ \midrule\bottomrule
	\end{tabular}
	}
	\label{table:kitti_comparison}
	\vspace{-0.15cm}
	\begin{flushleft}
	$\dagger$: Used for training \\
	$\mathsection$: A-LOAM + StickyPillars 
	
	\end{flushleft}
	\label{fig:kitti_odom}
	\vspace{-0.5cm}
\end{table}
\endgroup

\begingroup
\begin{table}[t!]
    \captionsetup{font=footnotesize}
	\centering
	\caption{Comparison in loop-closing situations with the state-of-the-art methods on Seq.~\texttt{06} of the KITTI dataset~\cite{geiger2012kitticvpr}. \textcolor{qwr}{The bold and the \textcolor{qwe}{gray-highlight} denote the best and the second-best performance, respectively} ($t_\text{avg}$:~m, $r_\text{avg}$:~$\deg$).}  
	\vspace{-0.15cm}
	\setlength{\tabcolsep}{4pt}
	{\scriptsize
	
	\begin{tabular}{l|cccccc}
	
	\toprule \midrule
	\multirow{2}[3]{*}{Method} & \multicolumn{2}{c}{0 $\sim$ 2m} & \multicolumn{2}{c}{4 $\sim$ 6m} & \multicolumn{2}{c}{8 $\sim$ 10m} \\  \cmidrule(lr){2-3} \cmidrule(lr){4-5} \cmidrule(lr){6-7} 
	 & $t_{\text{avg}}$ & $r_{\text{avg}}$ & $t_{\text{avg}}$ & $r_{\text{avg}}$ & $t_{\text{avg}}$  & $r_{\text{avg}}$   \\ \midrule
	 RANSAC10K~\cite{fischler1981ransac} & 2.369 & 14.22 & 5.010 & 27.35 & 8.341 & 33.34  \\
	 FGR~\cite{zhou2016fast}& \textbf{0.057} & \hl{0.222} & \hl{0.103} & \hl{0.301} & 1.821 & 1.828  \\
	 TEASER++~\cite{yang2020teaser}& 0.070 & 0.285 & 0.131 & 0.481 & 0.498 & 1.469  \\
	 Quatro (Ours) & 0.067 & 0.324 & 0.120 & {0.465} & \hl{0.471} & \hl{0.724}   \\ 
	 Quatro w/ INS (Ours) & \hl{0.059} & \textbf{0.207} & \textbf{0.101} & \textbf{0.230} & \textbf{0.429} & \textbf{0.346}   \\ 
	\midrule\bottomrule
	\end{tabular}
	}
	\label{table:kitti_IMU}
	\vspace{-0.3cm}
\end{table}
\endgroup
 
\subsection{Application: Leveraging INS in Non-flat Regions}

In addition, we conducted a feasibility study on the utilization of \textcolor{qwr}{an} INS system. That is, Quasi-SO(3) estimation is followed by \textcolor{qw}{the estimation of roll and pitch angles via INS measurements}, i.e. $\hat{\mathbf{R}}_y \cdot \hat{\mathbf{R}}_x$. 
As presented in Table~\ref{table:kitti_IMU}, even though the raw measurements were used, leveraging pitch and roll measurements reduced the rotation error effectively. Therefore, the results show the possibility of generalization of our proposed method in non-flat regions.

\begin{figure}[t!]
    \captionsetup{font=footnotesize}
	\centering 
	\begin{subfigure}[b]{0.44\textwidth}
		\includegraphics[width=1.0\textwidth]{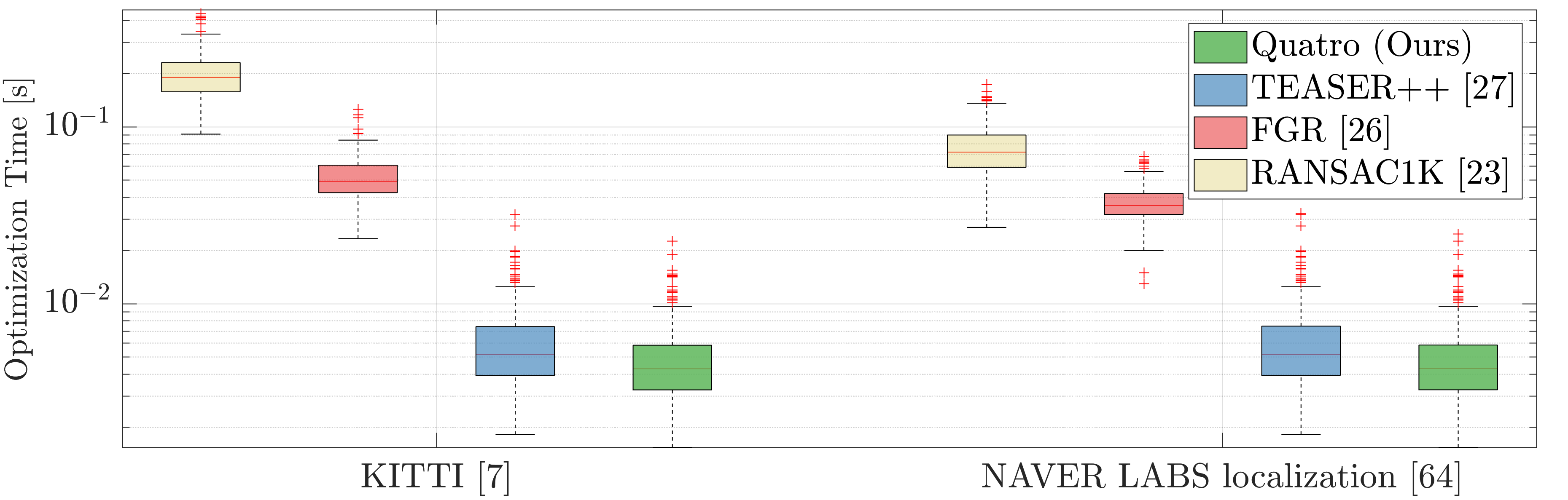}
	\end{subfigure}
	\vspace{-0.1cm}
	\caption{Average optimization time in the KITTI dataset~\cite{geiger2012kitticvpr} and the \textcolor{qw}{NAVER LABS localization} dataset~\cite{lee2021large_naverlabs} on Intel(R) Core(TM) i9-9900KF.} 
	\label{fig:optim_time}
	\vspace{-0.5cm}
\end{figure}

\subsection{Registration Speed}
 \vspace{-0.1cm}
 
Furthermore, our proposed method shows the fastest optimization time by virtue of MCIS-\texttt{heuristic}, as represented in Fig.~\ref{fig:optim_time}. On average, our method only takes 5~msec per optimization, which is sufficient for using our proposed method as a coarse alignment in real-time. 

\section{Conclusion}
 \vspace{-0.1cm}

In this study, a robust global registration method, \textit{Quatro}, has been proposed. Our proposed method proved to be more robust against degeneracy compared with the state-of-the-art methods. In future works, we plan to apply the concept of Quatro for generalization on various platforms, including UAVs, backpack-type mapping systems, and so forth.




\newpage

\bibliographystyle{IEEEtran}
\bibliography{./icra22,./IEEEabrv}

\end{document}